\pdfoutput=1
\documentclass[10pt,twocolumn,letterpaper]{article}

\usepackage{cvpr}              % To produce the CAMERA-READY version
\usepackage{amsmath,amssymb}
\usepackage{booktabs}
\usepackage{xcolor} 
\usepackage{colortbl}  
\usepackage{multirow}
\usepackage{multicol}
\usepackage{longtable} 
\usepackage{makecell}
\usepackage[symbol]{footmisc}
\usepackage[numbers,sort&compress]{natbib} 
\usepackage{algorithm}
\usepackage{algorithmic}
\usepackage[utf8]{inputenc}
\usepackage[T1]{fontenc}
\usepackage{array}
\usepackage{ragged2e}
\usepackage{cuted} 
\usepackage{supertabular}
\usepackage{enumitem}
\usepackage{afterpage}
\usepackage{listings}
\usepackage{microtype}
\usepackage{verbatim}
\usepackage{pifont} 
\usepackage[frozencache,cachedir=.]{minted}

\usepackage{tcolorbox}
\tcbuselibrary{minted, breakable}

\newcommand{\TODO}[1]{\textbf{\color{red}[TODO: #1]}}

\definecolor{darkgreen}{rgb}{0.0, 0.5, 0.0} % RGB values 
\definecolor{orange}{rgb}{1.0, 0.5, 0.0}   % Orange color

\newcommand{\redx}{\textcolor{red}{\ding{55}}}  % Red X
\newcommand{\greencheck}{\textcolor{darkgreen}{\checkmark}}  % Green checkmark
\newcommand{\orangecheck}{\textcolor{orange}{\checkmark}}  % Orange checkmark

\renewcommand{\TODO}[1]{}

\newcommand{\et}[2]{${#1}^{\pm{#2}}$}

\definecolor{cvprblue}{rgb}{0.21,0.49,0.74}
\usepackage[pagebackref,breaklinks,colorlinks,citecolor=cvprblue]{hyperref}

\usepackage[pagebackref,breaklinks,colorlinks]{hyperref}

\usepackage[capitalize]{cleveref}
\crefname{section}{Sec.}{Secs.}
\Crefname{section}{Section}{Sections}
\Crefname{table}{Table}{Tables}
\crefname{table}{Tab.}{Tabs.}

\def\framework{CoMA}
\def\model{SPAM}
\def\capmodel{MVC}

\begin{document}

\title{CoMA: Compositional Human Motion Generation with Multi-modal Agents}

\author{%
    Shanlin Sun$^1\footnotemark[1]$ \quad 
  Gabriel De Araujo$^1\footnotemark[1]$ \quad 
  Jiaqi Xu$^3\footnotemark[1]$ \quad 
  Shenghan Zhou$^4\footnotemark[1]$ \quad \\
  Hanwen Zhang$^5$ \quad 
  Ziheng Huang$^6$ \quad 
  Chenyu You$^2$ \quad 
  Xiaohui Xie$^1$\\~
\normalsize $^1$ University of California, Irvine \quad
$^2$ Stony Brook University \quad
$^3$ Southeast University \quad \\
$^4$ Chongqing University \quad 
\normalsize $^5$ Huazhong University of Science and Technology \quad
$^6$ Northeastern University 
}

\twocolumn[
    \begin{@twocolumnfalse}
        \maketitle 
        \vspace{-2em}
\begin{center}
    \centering
    \captionsetup{type=figure}
    \includegraphics[width=\linewidth]{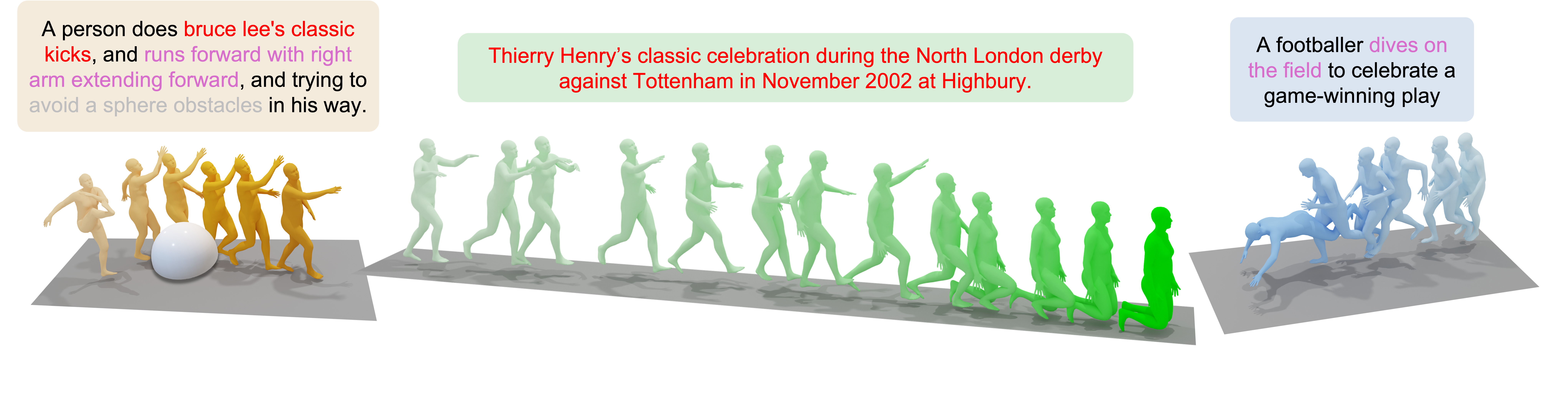}
    \vspace{-1.5em}
    \captionof{figure}{CoMA can generate high quality motion sequences despite challenging user expectations. Label colors \textcolor{red}{red} indicate context-rich moves and/or poses, \textcolor{Thistle}{purple} indicate spatially compositional motions and \textcolor{gray}{gray} indicate trajectory-editing instructions.}
    \label{fig:headline}
\end{center}
    \end{@twocolumnfalse}
]

\renewcommand{\thefootnote}{\fnsymbol{footnote}}
\footnotetext[1]{Equal contribution.}

%%%%%%%%% ABSTRACT
\begin{abstract}
3D human motion generation has seen substantial advancement in recent years. While state-of-the-art approaches have improved performance significantly, they still struggle with complex and detailed motions unseen in training data, largely due to the scarcity of motion datasets and the prohibitive cost of generating new training examples. To address these challenges, we introduce \textbf{\framework}, an agent-based solution for complex human motion generation, editing, and comprehension. \framework\ leverages multiple collaborative agents powered by large language and vision models, alongside a mask transformer-based motion generator featuring body part-specific encoders and codebooks for fine-grained control. Our framework enables generation of both short and long motion sequences with detailed instructions, text-guided motion editing, and self-correction for improved quality. Evaluations on the HumanML3D dataset demonstrate competitive performance against state-of-the-art methods. Additionally, we create a set of context-rich, compositional, and long text prompts, where user studies show our method significantly outperforms existing approaches. Project Page: \url{https://gabrie-l.github.io/coma-page/}
\end{abstract}

%%%%%%%%% BODY TEXT
\section{Introduction}
\label{sec:intro}

3D human motion generation has become increasingly vital across various applications, from gaming and virtual reality to robotics, spurring significant research interest. Among the emerging approaches, text-to-motion generation \cite{temos, t2m, motionclip, t2mgpt, tm2t, mdm, fgt2m, gmd, motiongpt, remodiffuse, motiondiffuse}, which leverages advances in natural language processing, faces distinct challenges. These challenges primarily stem from two factors: the limited availability of high-quality motion data due to costly acquisition processes, and the inherent complexity of mapping diverse possible motions to text descriptions.

Recent advances in generative approaches have significantly improved the state-of-the-art in this field. Diffusion models \cite{finemogen, motiondiffuse, remodiffuse, mld, mdm, fgt2m}, exemplified by MDM \cite{mdm}, excel in generating diverse motions but face challenges with fine-grained details and computational efficiency. Vector quantized VAE (VQ-VAE) \cite{vqvae} based methods \cite{mmm, momask, t2mgpt, bamm} address these limitations with recent masked transformer-based frameworks like MMM \cite{mmm} and MoMask \cite{momask}, achieving superior generation quality and inference speed. Building upon these foundational models, researchers have developed specialized motion editing methods conditioned on various inputs: text prompts \cite{motionfix, mmm}, trajectory keypoints \cite{gmd,omnicontrol}, joint locations \cite{gmd, dno, omnicontrol, priormdm}, and partial motions \cite{priormdm}.

\begin{figure}[ht]
    \centering
    \includegraphics[width=\columnwidth]{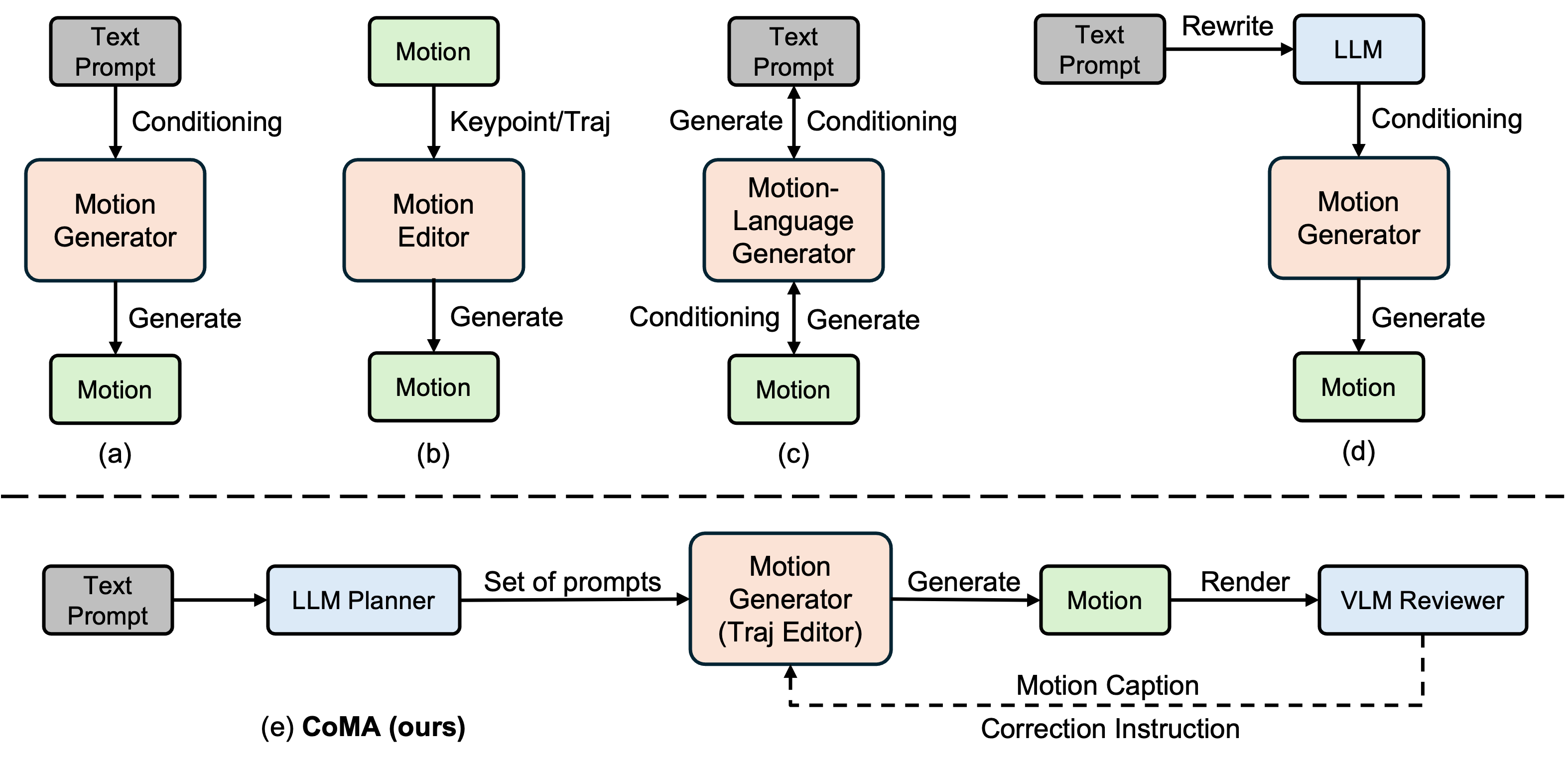}
    \caption{Illustrative architecture comparison between (a) text-conditional motion generation models~\cite{mdm,mld,momask,mmm}, (b) keypoint/trajectory-conditional motion editing models~\cite{gmd, dno, priormdm, motionfix, omnicontrol}, (c) Motion-language autoregressive models~\cite{motiongpt, motionchain, motionagent}, (e) LLM-grounded motion generation models~\cite{remodiffuse, finemogen, como} and (d) our CoMA framework.}
    \label{fig:coma_framework_comp}
\end{figure}
However, these methods show performance degradation when processing context-rich motion descriptions absent from training datasets. This limitation has led to the integration of Large Language Models (LLMs) for translating general user inputs into model-compatible prompts. Notable examples include FineMoGen \cite{finemogen} and CoMo \cite{como}, which developed approaches for body part-specific instructions, while MotionGPT \cite{motiongpt}, Motion-Agent \cite{motionagent} and MotionChain \cite{motionchain} explore conversational interfaces for generation and editing.

As illustrated in Fig.~\ref{fig:coma_framework_comp}, existing motion generation methods can be categorized into four main approaches. The first category comprises text-conditional motion generation models, encompassing both diffusion-based \cite{mdm, mld} and token modeling approaches \cite{momask, mmm, bamm}. The second category includes motion editing models that transform original motions, joint locations, or trajectories into new motions \cite{dno, gmd, omnicontrol, priormdm}. The third category consists of motion-language autoregressive models \cite{motionagent, motionchain, motiongpt} that integrate motion generation and understanding within unified multi-modal LLMs. The fourth category contains LLM-grounded motion generation models \cite{remodiffuse, finemogen, como} that utilize LLMs to parse user inputs into comprehensible prompts for motion generators.
Despite these advances, current motion generation methods still struggle with handling spatially and temporally compositional motions, even when individual body part movements and motion segments are manageable. This motivates us to propose \framework, a compositional human motion generation framework with multi-model agents.

As demonstrated in Fig.~\ref{fig:headline}, our framework successfully generates high-fidelity motions from complex inputs including long, context-rich descriptions, spatially compositional instructions, and trajectory-informed prompts. Tab.~\ref{tab:method_comp} highlights our framework's distinct advantages over recent motion generation methods. Compared to state-of-the-art approaches like MoMask \cite{momask} and MMM \cite{mmm}, \framework\ excels in handling complex and unseen user inputs through LLM-based prompt re-captioning. Unlike other LLM-grounded methods such as FineMoGen \cite{finemogen} and CoMo \cite{como}, our approach incorporates motion captioning capabilities, enabling self-correction. Moreover, in contrast to motion-language large generative models like MotionChain \cite{motionchain} and Motion-Agent \cite{motionagent}, \framework\ automatically decomposes complex motion tasks into manageable generation and editing sub-tasks.

The key contributions of our \framework\ framework (Fig.~\ref{fig:coma_framework_comp}e) include:
\begin{itemize}
    \item \textbf{Task Planner:} Leverages LLM's reasoning capabilities to decompose complex motion generation tasks into manageable sub-tasks and defines comprehensive generation pipelines, extending beyond simple user input translation.
    
    \item \textbf{Motion Generator:} Implements motion generation, editing, and sequence blending based on Task Planner instructions through our novel spatially-aware masked generative motion model (\model). This component demonstrates state-of-the-art performance on standard benchmarks and superior results for complex sequences in the HumanML3D \cite{humanml3d} dataset.
    
    \item \textbf{Trajectory Editor:} Provides optional trajectory manipulation, generating curve functions from textual descriptions and mapping keypoints along generated trajectories to motions.

    \item \textbf{Motion Reviewer:} Evaluates motion sequence fidelity against original text prompts. Our instruction-tuned video language model (\capmodel) demonstrates competitive performance on motion captioning tasks when evaluated on the HumanML3D dataset. Building upon this foundation, our Motion Reviewer agent effectively assesses motion-text alignment and generates correction instructions through LLMs.
    
    \item We introduce a challenging test prompt set demonstrating our pipeline's comprehensive handling of diverse text instructions, with user studies revealing significant advantages over existing state-of-the-art methods.
\end{itemize}

\begin{table}[t]
\resizebox{\columnwidth}{!}{%
\begin{tabular}{@{}lcccccc@{}}
\toprule
\multirow{2}{*}{Methods} & \multirow{2}{*}{\begin{tabular}[c]{@{}c@{}}Prompt\\Re-caption\end{tabular}} & \multirow{2}{*}{\begin{tabular}[c]{@{}c@{}}Motion\\Caption\end{tabular}} & \multicolumn{3}{c}{Composition} & \multirow{2}{*}{Self-correction} \\
\cmidrule(l){4-6}
& & & Spatial & Temporal & Task & \\
\midrule
MoMask \cite{momask}            & \redx & \redx   & \redx & \redx & \redx & \redx \\
MMM\cite{mmm}                   & \redx & \redx   & \redx & \redx & \redx & \redx \\
CoMo \cite{como}                & \greencheck & \redx   & \greencheck & \redx & \redx & \redx \\
\textit{FineMoGen}\cite{finemogen}       & \greencheck  & \redx   & \greencheck & \greencheck & \redx & \redx \\
Mandelli et al.\cite{italian}   & \greencheck  & \redx   & \greencheck & \greencheck & \redx & \redx \\
\textit{MotionChain}\cite{motionchain}  & \redx & \greencheck & \redx & \greencheck & \redx & \orangecheck \\
Motion-Agent \cite{motionagent} & \greencheck  & \greencheck & \redx & \greencheck & \redx & \orangecheck \\
\midrule
\framework~(Ours)               & \greencheck  & \greencheck & \greencheck & \greencheck & \greencheck & \greencheck \\
\bottomrule
\end{tabular}%
}
\caption{Comparison of recent state-of-the-art methods on diverse motion-relevant tasks. \greencheck~indicates full inclusion of the feature, \redx~indicates absence, and \orangecheck~indicates incompleteness. We deem \cite{motionagent,motionchain} to have incomplete self-correction capabilities as they need human-provided correction instructions. Italicized model
indicates the corresponding model requires additional human-annotated data for training.}
\label{tab:method_comp}
\end{table}
\section{Related Works}

\subsection{Text-driven Human Motion Generation}

The field of human motion generation has seen significant progress over recent years. Available through many modalities, such as text prompt conditioning, action label conditioning, or constraint free inputs, motion sequences composed of joint locations and their respective rotations have been addressed through a variety of methods. On the domain of text-conditioned motion generation, initial works \cite{l2p, plappert2018learning,ghosh2021synthesis,lin2018generating} proposed deterministic modeling approaches, leading to blurry generated sequences. This issue was posteriorly addressed through the advent of stochastic models. Subsequent works sought to explore VAE-based methods given their proved success in other generative tasks \cite{vae1,vae2,vae3}. T2M \cite{t2m} adopted such architecture to learn a probabilistic text to motion mapping, while TEMOS \cite{temos} and TEACH \cite{teach} leveraged  transformer-based VAEs to create a joint latent embedding of natural language and motion. Currently, both diffusion-based and autoregressive approaches have presented significant performance gains, and have rapidly taken the lead in the field both adoption and performance-wise. Diffusion models \cite{finemogen,mld,mdm, fgt2m, mld, motiondiffuse} emerged as powerful tools given their offered diversity in distributions and continuous representation of motions, leading to smooth and varied generations. Autoregressive-based works adopting Vector-Quantized Variational Autoencoders (VQ-VAEs) \cite{vqvae}, notably MotionGPT \cite{motiongpt}, MMM \cite{mmm} and MoMask \cite{momask}, highlight the efficiency in representing motions as discrete tokens and combining such data with an autoregressive transformer architecture for producing coherent and smooth sequences. 

While the above mentioned approaches offer an array of motion generation tasks, our work emphasizes complex, body part specific generations, while also offering longer, compositional and context-rich generations, text-based editing, and the ability to understand and correct its own generations if necessary.

\subsection{LLM-Integrated Motion Generation}
While text-driven human motion generation methods have achieved impressive results, they often struggle with uncommon text prompts. To address this limitation, several approaches have emerged that leverage Large Language Models (LLMs). ReModiffuse \cite{remodiffuse} pioneered this direction by integrating a retrieval mechanism into a diffusion-based framework to refine the denoising process.

With the recent popularization of LLMs, researchers have explored various ways to enhance motion generation without increasing model size. CoMo \cite{como} decomposes motions into discrete, semantically meaningful pose codes for each body part, enabling direct LLM-guided motion editing through code adjustment. FineMoGen \cite{finemogen} implements both spatial and temporal motion decomposition, supported by their fine-grained HuMMan-MoGen dataset that provides detailed body part annotations across multiple motion stages. Similarly, Mandelli et al. \cite{italian} propose using LLMs to break down complex actions into simpler, training-observed movements.

Another line of research focuses on unified motion-language models. MotionGPT \cite{motiongpt} treats motion as a language, creating a unified model for various motion-related tasks. MotionChain \cite{motionchain} extends this approach by supporting multi-turn interactions and image prompts through synthetic conversational data. Motion-Agent \cite{motionagent} takes a different approach by developing MotionLLM, a generative agent that bridges the motion-text gap. By integrating MotionLLM with GPT4 without additional training, it achieves complex motion generation through multi-turn conversations. However, both MotionChain and Motion-Agent require recurrent human interaction to fully utilize their reasoning capabilities.

In contrast, our \framework\ framework uniquely combines multi-modal agents to enable automatic iterative motion corrections. It unifies motion generation and fine-grained editing while maintaining compatibility with any LLM, VLM, and motion generative models. Notably, \framework\ achieves this without requiring additional training or data beyond the HumanML3D dataset \cite{humanml3d}.
\section{CoMA Overview}

\begin{figure}[ht!]
    \vspace{-1em}
    \centering
    \includegraphics[width=\columnwidth]{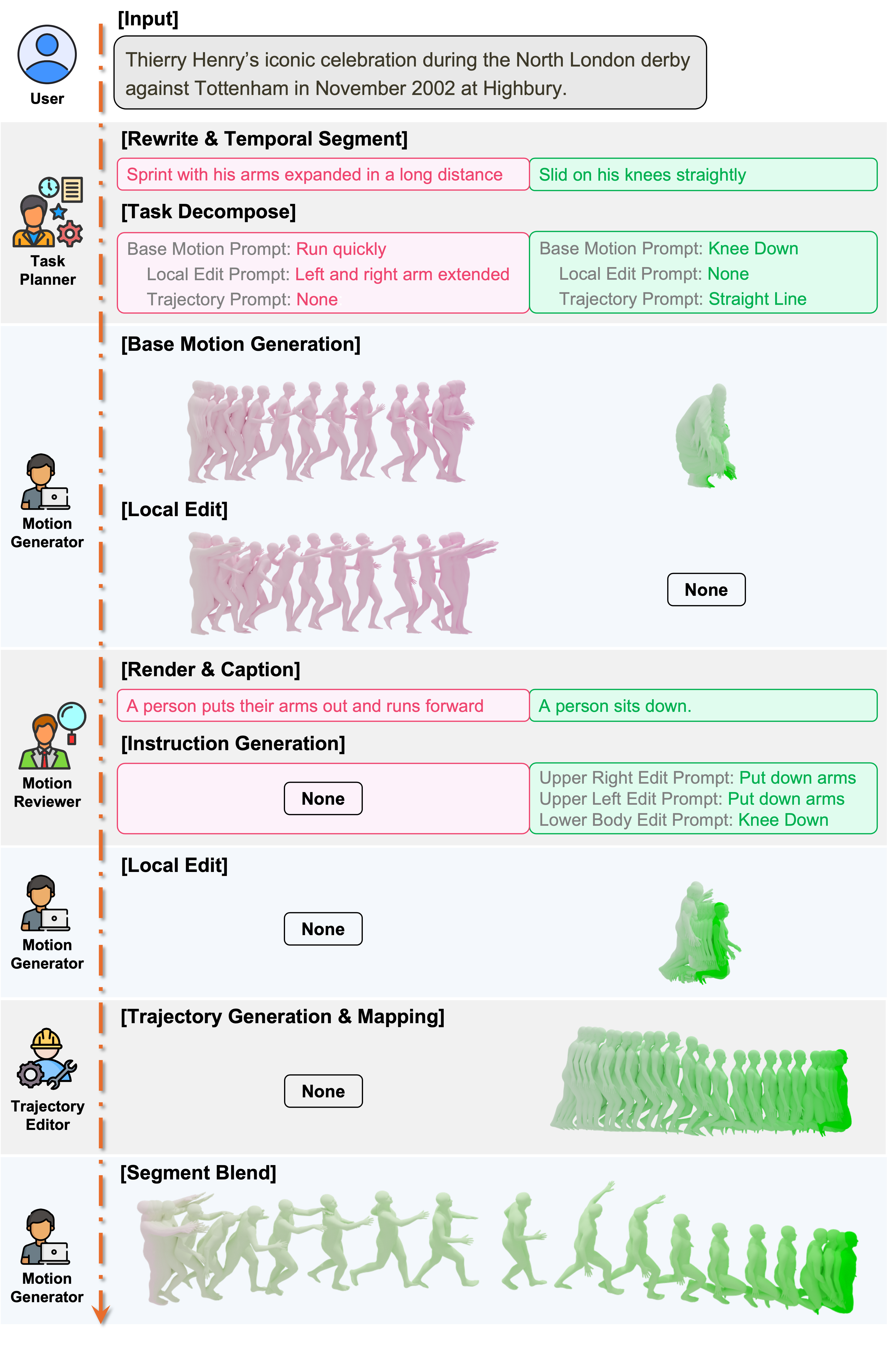}
    \caption{A real example of how our CoMA workflow generates context-rich, compositional and long motion sequence given only text prompt. More detailed explanations on this example are in Appendix.~\ref{app_subsec:zoom_in_example}}
    \label{fig:coma_workflow_example}
    \vspace{-1em}
\end{figure}

\framework~takes input as abstract and/or complex textual motion description and generates human motion sequences in a compositional manner; see Fig.~\ref{fig:coma_workflow_example}. To achieve this, we design a series of collaborative multi-modal agents to decompose the process of generating human motions into simpler, singular generation tasks in different temporal segments. Furthermore, we unify human motion generation and editing in an iterative closed-loop fashion. 

\subsection{Agent Functionality}
\label{subsec:agent_func}

Agents in \framework~can be categorized into a high-level task planner and several low-level actors. In the following subsections, we will introduce each agent's functionality during inference. We use GPT-4o \cite{gpt4} and VideoChat2 \cite{videochat2} for our and LLM and VLM models, respectively.

\noindent\textbf{Task Planner} reasons the input text prompt in three steps: text recaption, temporal segments, and task decomposition. 
First, we prompt GPT-4o to rewrite prompts to eliminate descriptions not contained in the motion datasets (such as falling on a tatami, performing the wakanda forever salute, etc), which may lead to the failure in the motion generation process. we let GPT-4o replace such textual abstractions (tatami $\rightarrow$ ground) and extract hidden motion information (wakanda forever salute $\rightarrow$ cross arms in the front of chest) from the original user input. The rewritten textual input is composed of terminologies retrieved from the training dataset, being better understood by the motion generation model. Thanks to the powerful generalization and reasoning ability of LLMs, few information is lost in this process. %

Secondly, we prompt GPT-4o to split the rewritten text into temporally consecutive segments. Two factors can negatively affect the performance of models to deal with longer text prompts: the CLIP encoder 77 token enconding limit, and the lack of long motion sequences in the training dataset. In response, we further split the rewritten text into temporally consecutive segments, each attributing a shorter but complete motion.

\begin{algorithm}[!ht]
\caption{Agent Collaboration Workflow in \framework.}
\label{alg:workflow}
\begin{algorithmic}[1]
\REQUIRE User text prompt $P$, Maximum number of self-correction round $K$
\STATE $P_{\text{concrete}} \leftarrow \text{TaskPlanner.}\texttt{Rewrite}(P)$
\STATE $\{P_i\}_{i=1}^N \leftarrow \text{TaskPlanner.}\texttt{Segment}(P_{\text{concrete}})$
\FOR{each segment $i$}
    \STATE $P_{\text{base}}^i, P_{\text{edit}}^i, P_{\text{traj}}^i \leftarrow \text{TaskPlanner.}\texttt{Decompose}(P_i)$
    \STATE $M_{\text{base}}^i \leftarrow \text{MotionGenerator.}\texttt{Generate}(P_{\text{base}}^i)$
    \STATE $M_i \leftarrow \text{MotionGenerator.}\texttt{Edit}(M_{\text{base}}^i, P_{\text{edit}}^i)$
    \FOR{$k \leftarrow 1$ to $K$}
        \STATE $V_i \leftarrow \text{MotionReviewer.}\texttt{Render}(M_i)$
        \STATE $C_i \leftarrow \text{MotionReviewer.}\texttt{Caption}(V_i)$
        \STATE $I_i \leftarrow \text{MotionReviewer.}\texttt{Instruct}(C_i, P_i)$
        \IF{$I_i == \emptyset$}
            \STATE \textbf{break}
        \ELSE
            \STATE $M_i \leftarrow \text{MotionGenerator.}\texttt{Edit}(M_i, I_i)$
        \ENDIF
    \ENDFOR
    \IF{$P_{\text{traj}}^i \neq \emptyset$}
        \STATE $T_i ~~\leftarrow \text{TrajectoryEditor.}\texttt{Generate}(P_{\text{traj}}^i)$
        \STATE $M_i ~~\leftarrow \text{TrajectoryEditor.}\texttt{Map}(M_i, T_i)$
    \ENDIF
\ENDFOR
\STATE $M_{\text{final}} \leftarrow \text{MotionGenerator.}\texttt{Blend}(\{M_i\}_{i=1}^N)$
\ENSURE Final motion sequence $M_{\text{final}}$
\end{algorithmic}
\end{algorithm}
Lastly, for each segment, GPT-4o decomposes a motion generation task into a base generation and local editing tasks. State-of-the-art motion generation models struggle with spatially compositional motions, such as ``walk while raising the left hand and lowering the right hand at the same time", and even if generating such motion is possible, it remains challenging to ensure the correctness of local details. Thus, we decompose the motion generation task to generate a global motion and local body motions separately. If trajectory information is available, we also require GPT-4o to extract it from the prompt and forward this geo-spatial description to the trajectory modification agent. Furthermore, our task planner can also use GPT-4o to estimate the duration of each segment, which is crucial to generate motions with realistic speed. The integration of the task planning agent into this system enhances its robustness in interpreting various text prompts and streamlines operations for clarity and fine granularity. More prompting details are in the Appendix~\ref{app_sec:task_planner}.

\noindent\textbf{Motion Generator}
unifies text-driven global human motion generation and local body part editing.
To this end, we propose \model, a masked generative model where four codebooks and encoders are learned to represent four body parts, while a shared motion decoder learns to output whole human motions by fusing four local body part codes. More details are in Sec.~\ref{sec:model}.

\noindent\textbf{Trajectory Editor}
is responsible for modifying the motion’s trajectory based on textual trajectory descriptions. 
By employing Chain-of-Thought (CoT) \cite{cot} reasoning to trigger GPT-4o’s spatial understanding, this agent generates various curve functions to produce accurate pelvis trajectories. Sampling, interpolation, and resampling subsequently yield precise key points, enabling reconstruction of rotation data. See more details in the Appendix~\ref{app_sec:trajectory_editor}.

\noindent\textbf{Motion Reviewer} evaluates whether a generated motion sequence faithfully represents the user's original text prompt by leveraging a Vision-Language Model (VLM).
If the generated motion is not aligned with text prompt, the Motion Reviewer generates specific correction instructions and returns the sequence to the Motion Generator for refinement.
We instruction-tune VideoChat2 on the HumanML3D dataset to enable accurate captioning of rendered motion sequences. The generated captions are then compared with the original text prompt using GPT-4o to generate correction prompts when necessary. Detailed implementation is provided in the Appendix~\ref{app_sec:motion_reviewer}.

\subsection{Agent Collaboration Workflow}
\label{subsec:workflow}
\framework~operates through a systematic multi-stage workflow that orchestrates the collaboration between different agents, as outlined in Algorithm.1. 
Given a user text prompt $P$, the Task Planner first processes it through three sequential steps: (1) prompt rewriting, where abstract concepts are transformed into concrete motion descriptions $P_{\text{concrete}}$; (2) temporal segmentation, where the rewritten prompt is divided into temporally consecutive segments; and (3) task decomposition, where each motion segment description is further decomposed into base motion prompt $P_{\text{base}}^i$ and local motion editing prompt $P_{\text{edit}}^i$, along with a trajectory prompt $P_{\text{traj}}^i$. For each segment, the Motion Generator first creates base motion sequences $M_{\text{base}}^i$ and applies local body part editing to get initial generated motion $M_i$. To ensure motion quality, we implement an iterative self-correction loop where the Motion Reviewer evaluates the generated motions by comparing motion video rendering caption $C_i$ with input text prompt $P_i$. The Motion Reviewer generates refinement instructions $I_i$ if there is significant discrepancy between $C_i$ and $P_i$, and the process returns to the editing stage. This correction loop continues until either no further instructions are needed (i.e., $I_i == \emptyset$) or it reaches the maximum number of iterations $K$. If trajectory control is required (i.e., $P_{\text{traj}}^i \neq \emptyset$), the Trajectory Editor creates precise 2D motion trajectories ($T_i$) and maps them to the generated motion segment $M_i$. Finally, the Motion Generator blends all segment sequences ($\{M_i\}_{i=1}^N$) to produce the complete motion sequence ($M_{\text{final}}$).

\section{Motion Generator}
\label{sec:model}
\subsection{Preliminary Knowledge}
\label{subsec:preliminary}

MMM~\cite{mmm} transforms motion sequences into discrete tokens using VQVAE~\cite{vqvae}. Given a motion sequence $\mathbf{m}_{1:N} \in \mathbb{R}^{N\times D}$, a 1D convolutional encoder $\mathcal{E}$ first encodes it into latent vectors $\mathbf{b}_{1:n} \in \mathbb{R}^{n \times d}$ with downsampling ratio $n/N$. Each vector is then quantized to its nearest neighbor from a codebook $\mathcal{C} = \{c_k\}_{k=1}^{K} \subset \mathbb{R}^d$ via $\mathcal{Q}(\cdot)$, producing $\mathbf{\tilde b}_{1:n} = \mathcal{Q}(\mathbf{b}_{1:n})$. A decoder $\mathcal{D}$ reconstructs the motion as $\mathbf{\tilde m} = \mathcal{D}(\mathbf{\tilde b})$, with the codebook indices serving as discrete motion tokens.
MoMask~\cite{momask} extends this using residual vector quantization (RVQ)~\cite{soundstream} to produce multiple token layers. Starting with $r_0 = \mathbf{b}$, each layer $v$ recursively computes:

\begin{equation}
\mathbf{\tilde b}^v = \mathrm{Q}(\mathbf{r}^v), \quad \mathbf{r}^{v+1} = \mathbf{r}^v - \mathbf{\tilde b}^v
\end{equation}

where $v = 0, \dots, V$. The final latent approximation $\sum_{v=0}^{V} \mathbf{ \tilde b}^{v}$ is then decoded through $\mathcal{D}$.

For text-guided generation, MoMask uses two transformers: a masked transformer generating base-layer tokens $t_{1:n}^0$ with masking schedule $\gamma(\tau)=\cos(\frac{\pi\tau}{2})$, and a residual transformer sequentially predicting tokens for layers $1$ to $V$. Both employ classifier-free guidance during inference, computing logits as $\omega_g = (1+s)\cdot\omega_c - s\cdot\omega_u$, where $\omega_c$ and $\omega_u$ are conditional and unconditional predictions.
\subsection{\model}
\begin{figure*}[ht]
    \vspace{-2em}
    \centering
    \includegraphics[width=0.95\textwidth]{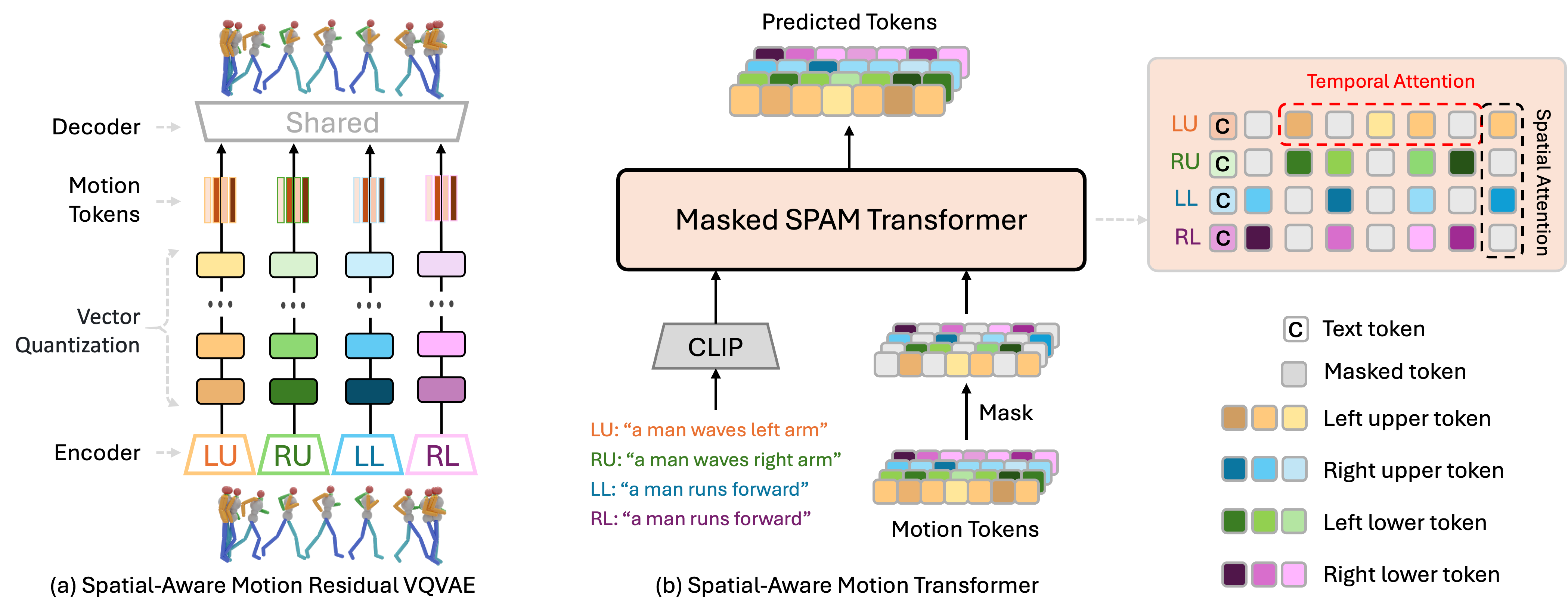}
    \caption{\textbf{SPAM overview.} (a) Motion sequence is decomposed into four body parts: left upper (LU), right upper (RU), left lower (LL), and right lower (RL). Each part is tokenized through separate RVQs and reconstructed into a whole-body motion through a shared decoder. (b) Base-layer motion tokens are randomly masked, while local/global text prompts are encoded separately and concatenated with corresponding motion tokens. The Masked SPAM Transformer is trained to predict the masked tokens. The residual transformer follows a similar architecture and is omitted for brevity.}
\label{fig:spam}
\end{figure*}
\framework~aims to deliver a unified Motion Generator agent that not only generates complex human motions from global text prompts in one shot, but also understands granular editing instructions to modify specified body parts.
We propose a Spatially-Aware Masked Generative Motion Model (\model), which processes both local and global text prompts to coherently generate and/or edit four body parts (right/left upper/lower, see Appendix~\ref{app_subsec:body_parts} for details).
Given that our method builds upon MoMask, we focus on explaining the key differences between our model and the original MoMask architecture.

\subsubsection{Spatially-Aware Motion Residual VQVAE}
\label{subsec:vqvae}
Our spatially-aware VQVAE consists of four encoders, four separate quantizers and one shared decoder, as is shown in Fig.~\ref{fig:spam}(a). 
Each body part has one codebook and its own encoder, which converts the corresponding motion sequence $ \mathbf{m}^{i} \in \mathbb{R}^{N \times D_{i}} $ into a latent vector sequence $ \mathbf{b}_{i} \in \mathbb{R}^{n \times d} $. Thus, one motion sequence can be represented by four tuples of body parts motion tokens:
\begin{equation}
\mathbf{{B}} = \left[ \mathbf{{b}}_{i} \right]_{i=1}^4 = \in \mathbb{R}^{4 \times n \times d}
\end{equation}
each of which is generated by a corresponding quantizer and encoder. Finally, the four body parts motion tokens are concatenated, which will be decoded into motion space by a shared decoder, generating whole body motions:
\begin{equation}
\mathbf{\tilde{m}}= \mathcal{\hat{D}} \left( \text{concat} \left( \left[ \mathcal{Q}_{i} \left( \mathcal{E}_{i} \left( \mathbf{m}^{i} \right) \right) \right]_{i=1}^{4} \right) \right)
\end{equation}

Following MoMask, we train the residual motion VQ-VAEs via a motion reconstruction loss combined with a latent embedding loss at each quantization layer:
\begin{equation}
\mathcal{L}_{\text{rvq}} = \| \mathbf{m} - \tilde{\mathbf{m}} \|_1 + \beta \sum_{v=1}^{V} \| \mathbf{R}^v - \text{sg}[\mathbf{B}^v] \|_2^2
\end{equation}
where $\mathbf{R}^v = [r^{v}_{i}]_{i=1}^{4}$ denotes the residual tokens tuples, $ \text{sg}[\cdot] $ denotes the stop-gradient operation, and $ \beta $ is a weighting factor for the embedding constraint. This framework is optimized with a straight-through gradient estimator \cite{vandenoord2017neural}, and our codebooks are updated via exponential moving average and codebook reset following T2M-GPT \cite{t2mgpt}.

\subsubsection{Spatially-Aware Motion Transformer}
\label{subsec:trans}

Our Spatially-Aware Transformer models both base-layer motion token tuples $ T^{0} = [ t_{i}^{0} ]_{i=0}^{4} \in \mathbb{R}^{4 \times n} $ and residual-layer motion token tuples $\left [ T^{v} \right ]_{v=1}^{V} \in \mathbb{R}^{V\times 4 \times n}$ using base transformer $ f_\theta $ and residual transformer $ f_\phi $, where each base token tuple $ t_{i}^{0}$ represents a distinct body part.
Inspired by video classification architectures~\cite{gberta_2021_ICML}, we implement a factorized space-time self-attention mechanism for motion generation. As is shown in ~\cref{fig:spam}(b), our \model~transformer splits the attention computation into two sequential steps: spatial attention across body parts, followed by temporal attention across time steps.
In this design, tokens first attend to others within the same time step through spatial self-attention, capturing inter-part relationships. Subsequently, temporal self-attention is applied to each spatial position across time steps to model temporal dependencies. This factorized approach reduces computational complexity while maintaining expressiveness by separately modeling spatial and temporal relationships.

\noindent\textbf{Base Transformer} As shown in the ~\cref{fig:spam}(b), given masked motion tuples $ \hat{T}^0 $ and text prompts $ P = [\mathbf{p}^{i}]_{i=0}^{4} $, where $\mathbf{p}^{i}$ describes body part $i$, the base transformer predicts the masked tokens. The text prompts can be either identical global descriptions or distinct part-specific instructions. We extract text features using CLIP~\cite{motionclip}. The base transformer $ f_\theta $ is trained to minimize:
\begin{equation}
\mathcal{L}_{\text{base}} = \sum_{\hat{T}_k = [\text{MASK}]} -\log f_\theta(T_k^0 \mid \hat{T}^0, P)
\end{equation}

\noindent\textbf{Residual Transformer} The residual transformer $f_\phi$ mirrors the base transformer's architecture but maintains $V$ separate embedding layers. Given a randomly selected layer $j \in [1, V]$, it embeds and sums tokens from preceding layers $T^{0:j-1}$, then predicts tokens for layer $j$ conditioned on these embeddings, text $P$, and layer index $j$. The training objective is:
\begin{equation}
\mathcal{L}_{\text{res}} = \sum_{j=1}^{V} \sum_{i=1}^{n} -\log f_\phi(T_i^j \mid T^{0:j-1}_i, P, j)
\end{equation}

\subsubsection{Motion Editing}
\label{subsec:edit}
For complex motions, SPAM sometimes struggles to generate satisfactory results in a single attempt, which require the composition of simple generation and fine editing, which is collaborated within our \framework\ system.
\model~supports multiple motion editing tasks to iteratively refine the motion. All of the editing tasks below do not require additional training and can seamlessly integrate with each other.

\noindent\textbf{In-between Editing} After the user sets frames $\mathbf{\alpha:\beta}$ to be edited, corresponding token tuples $\mathbf{T}^{0}_{\alpha:\beta}$ will be replaced with [MASK]. Our SPAM will fill in these [MASK] tokens and generate a natural animation. Motion Reviewer agent can automatically select the keyframes to be in-paint / re-paint.

\noindent\textbf{Body Part Editing} Our model supports text-driven editing of four body parts: left upper, right upper, left lower, and right lower. After specifying the body parts $J$ to be edited, the corresponding token sequences $[\mathbf{t}_{j}^{0}]_{j \in J}$ will be replaced with \texttt{[MASK]}. To ensure a natural connection between the other body parts and the edited parts, we introduce random \texttt{[MASK]} tokens into the other body parts. The Motion Reviewer agent will automatically select the parts to be edited until the desired result is achieved.

\noindent\textbf{Blend Editing} Inspired by MMM, given a sequence of motions, the model will generate transition motion tokens conditioned on the end of the previous motion sequence and the start.

%------------------------------------------------------------------------------------------------------------------------------------------------
\section{Experiments}
We evaluated \framework~from two perspectives: quantitative performance on the standard HumanML3D benchmark \cite{humanml3d}, and qualitative assessment through human studies focused on complex motion generation. 

\subsection{Experiments on HumanML3D}
\subsubsection{Setup}
HumanML3D contains 14,616 motions extracted from multiple source datasets, with each motion paired with 3 textual descriptions, totaling 44,970 possible prompts. We use the standard split comprising 23,384 training samples, 1,460 validation samples, and 4,384 test samples.

Following prior works in motion generation \cite{mdm, momask, t2m, mmm, motiongpt, como, temos, motiondiffuse, remodiffuse, t2mgpt, mld}, we adopt standard evaluation metrics: Frechet Inception Distance (FID) for measuring distributional similarity between generated and ground truth motions, R-Precision and Multimodal Distance (matching score) for assessing text-motion semantic alignment, and Multimodality for quantifying generation diversity.

We compare our method against state-of-the-art approaches across three categories: diffusion-based methods (MDM \cite{mdm}, FineMoGen \cite{finemogen} and CoMo \cite{como}), masked generation methods (MMM \cite{mmm} and MoMask \cite{momask}), and autoregressive methods (T2M-GPT \cite{t2mgpt} and MotionGPT \cite{motiongpt}), as well as large motion-language models (MotionChain \cite{motionchain} and Motion-Agent \cite{motionagent}).

\subsubsection{Generation Results}
\newcommand{\best}[2]{
    \cellcolor{red!25}\et{#1}{#2}
}

\newcommand{\secondbest}[2]{
    \cellcolor{orange!25}{\et{#1}{#2}}
}

\newcommand{\thirdbest}[2]{
    \cellcolor{yellow!25}{\et{#1}{#2}}
}

\begin{table*}[thb]
    \centering
    \scalebox{0.9}{
    \begin{tabular}{l c c c c c c}
    \toprule
    \multirow{2}{*}{Methods}  & \multicolumn{3}{c}{R Precision$\uparrow$} & \multirow{2}{*}{FID$\downarrow$} & \multirow{2}{*}{MultiModal Dist$\downarrow$} & \multirow{2}{*}{MultiModality$\uparrow$}\\

    \cline{2-4}
    ~ & Top 1 & Top 2 & Top 3 \\
    \midrule
        MDM \cite{mdm}& \et{0.320}{.005} & \et{0.498}{.004} & \et{0.611}{.007} & \et{0.544}{.044} & \et{5.566}{.027} & \best{2.799}{.072}  \\
        T2M-GPT \cite{t2mgpt}& \et{0.492}{.003} & \et{0.679}{.002} & \et{0.775}{.002} & \et{0.141}{.005} & \et{3.121}{.009}  & \et{1.831}{.048}  \\
        CoMo \cite{como}& \et{0.502}{.002} & \et{0.692}{.007} & \et{0.790}{.002} & \et{0.262}{.004} & \et{3.032}{0.015} & \et{1.013}{.046}  \\
        FineMoGen \cite{finemogen}& \et{0.504}{.002} & \et{0.690}{.002} & \et{0.784}{.002} & \et{0.151}{.008} & \et{2.998}{.008} & \secondbest{2.696}{.079}  \\
        MotionChain \cite{motionchain}& \et{0.504}{.003} & \et{0.695}{.003} & \et{0.790}{.003} & \et{0.248}{.009} & \et{3.033}{.010} & \et{1.715}{.066}  \\
        Motion-Agent \cite{motionagent}& \thirdbest{0.515}{.004} & - & \thirdbest{0.801}{.004} & \et{0.230}{.009} & \et{2.967}{.020} & -  \\
        MotionGPT \cite{motiongpt}& \et{0.492}{.003} & \et{0.681}{.003} & \et{0.778}{.002} & \et{0.232}{.008} & \et{3.096}{.008} & \thirdbest{2.008}{.084}  \\
        MMM \cite{mmm}& \thirdbest{0.515}{.002} & \secondbest{0.708}{.002} & \secondbest{0.804}{.002} & \secondbest{0.089}{.005} & \best{2.926}{.007} & \et{1.226}{.035}  \\
        MoMask \cite{momask}& \secondbest{0.521}{.002} & \best{0.713}{.002} & \best{0.807}{.002} & \best{0.045}{.002} & \thirdbest{2.958}{.008} & \et{1.241}{.040}  \\
        
        \toprule      
        \model & \best{0.526}{.003} & \best{0.713}{.003} & \secondbest{0.804}{.002} & \thirdbest{0.108}{.004} & \secondbest{2.939}{.008} & \et{0.924}{.039}  \\
    \bottomrule
    \end{tabular}
    }
    \caption{\textbf{Quantitative evaluation on the HumanML3D test set.} $\pm$ indicates a 95\% confidence interval. \textbf{\textcolor{red}{red}}, \textbf{\textcolor{orange}{orange}}, \textbf{\textcolor{yellow}{yellow}} indicates first, second and third best results, respectively.}
    \label{tab:quantitative_eval}
\end{table*}
Tab.~\ref{tab:quantitative_eval} presents results on the standard HumanML3D benchmark. Our \model\ achieves top-3 performance in FID, Multimodal Distance, and R-Precision metrics. Notably, we achieve the best performance in Top-1 and Top-2 R-precision while ranking second in Top-3, demonstrating our model's superior instruction-following capability.

\subsection{Results on Fine-grained Text Prompts}
\begin{table}[thb]
    \centering
    \scalebox{0.8}{ % Adjust the scaling factor as needed
        \begin{tabular}{l c c c c}
        \toprule
        \multirow{2}{*}{Methods} & \multicolumn{3}{c}{R Precision$\uparrow$} & \multirow{2}{*}{FID$\downarrow$} \\
        \cline{2-4}
        ~ & Top 1 & Top 2 & Top 3 \\
        \midrule
        MMM & \secondbest{0.446}{.004} & \secondbest{0.620}{.003} & \secondbest{0.716}{.003} & \secondbest{0.470}{.011} \\
        MoMask & \thirdbest{0.435}{.003} & \thirdbest{0.613}{.003} & \thirdbest{0.711}{.002} & \thirdbest{0.667}{.017} \\
        \midrule
        \model & \best{0.488}{.005} & \best{0.674}{.005} & \best{0.771}{.003} & \best{0.208}{.015} \\
        \bottomrule
        \end{tabular}
    }
    \caption{Quantitative evaluation on the test split of HumanML3D, with fine-grained text descriptions generated by GPT-4o}
    \label{tab:zero_quantitative_eval}
\end{table}

Leveraging SPAM's spatial understanding of human body dynamics, \model\ demonstrates enhanced comprehension of fine-grained text descriptions generated by GPT-4. We evaluated this capability by comparing SPAM with leading models like MoMask and MMM on the HumanML3D test set using GPT-4-enhanced descriptions. These descriptions augment the original ground truth texts with more detailed motion specifications.

As shown in Tab.~\ref{tab:zero_quantitative_eval}, while enhanced descriptions provide richer details, they also demand stronger spatial understanding of human body mechanics. Traditional models like MoMask and MMM show significant performance degradation with these detailed prompts, lacking the necessary spatial comprehension capabilities.

\subsection{Editing Capabilities}
\begin{figure}[ht]
    \vspace{-1em}  % Decrease space before the figure
    \centering
    \includegraphics[width=\columnwidth]{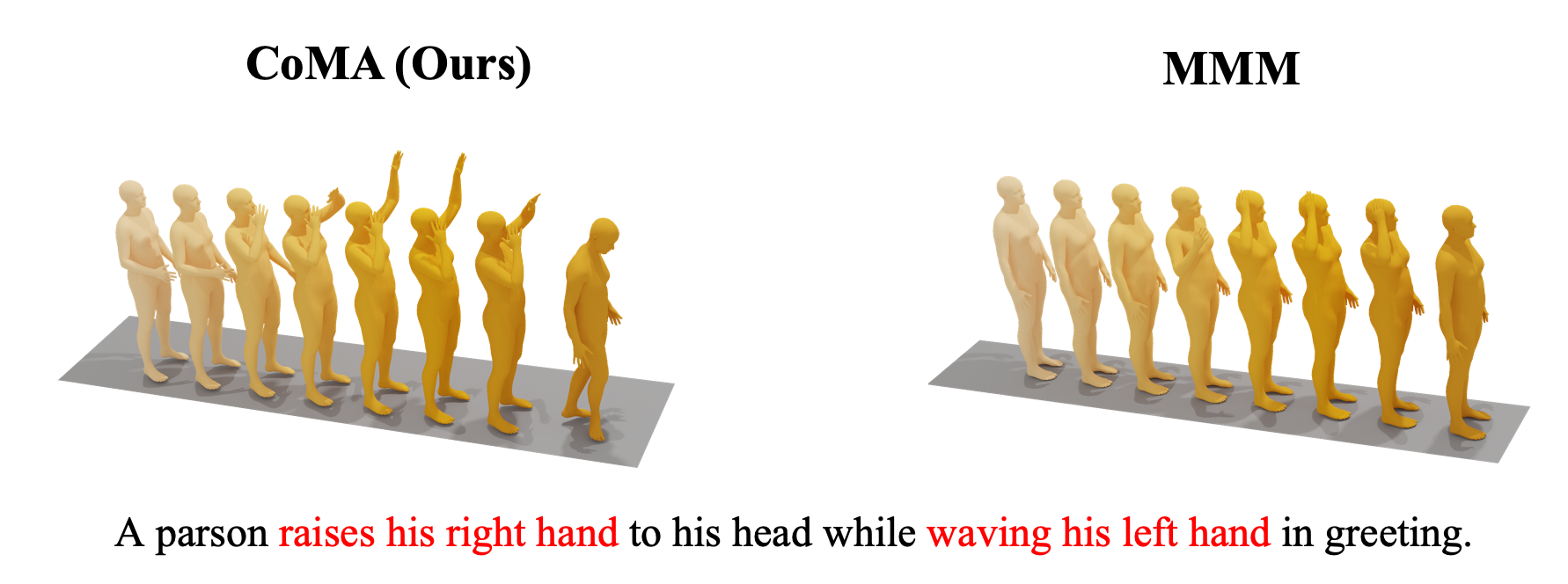}
    \vspace{-2em}  % Decrease space after the figure and caption
    \caption{Editing abilities of CoMA and MMM}
    \label{fig:edit}
\end{figure}

SPAM's spatial understanding enables precise motion editing across four main body parts. While existing methods like MMM \cite{mmm} support basic upper/lower body division, they struggle with fine-grained editing tasks. Fig.~\ref{fig:edit} demonstrates this limitation: given the input "A person raises his right hand to his head," when editing to include "while waving his left hand in greeting," MMM fails to preserve the original right-hand motion. In contrast, \framework~not only allows specific body part editing but also provides automatic modification suggestions through VLM integration.
\subsection{Motion Caption Results}

\newcommand{\bestmas}[1]{\cellcolor{red!25}#1}
\newcommand{\secondmas}[1]{\cellcolor{orange!25}#1}
\newcommand{\thirdmas}[1]{\cellcolor{yellow!25}#1}

\begin{table}[thb]
\centering
\scalebox{0.75}{\begin{tabular}{lccccc}
\toprule
Model & Bleu@1 $\uparrow$& Bleu@4 $\uparrow$& Rouge $\uparrow$& Cider $\uparrow$& Bert Score $\uparrow$ \\
\midrule
TM2T        & \thirdmas{48.90}  & 8.27     & \thirdmas{38.1}  & 15.80     & 32.2      \\
MotionGPT   & 48.20  & 12.47    & 37.4  & 29.20     &  32.4      \\
MotionChain & 48.10  & \thirdmas{12.56}    & 33.9  & \thirdmas{33.70}     & \thirdmas{36.9}      \\
MotionLLM   & \secondmas{54.53} & \secondmas{17.65}    & \bestmas{48.7}  & \secondmas{33.74}    & \bestmas{42.6}     \\
\midrule
\textbf{MVC (ours)}  & \bestmas{60.05} & \bestmas{20.98}  & \secondmas{45.79} & \bestmas{44.03}  & \secondmas{40.12}   \\
\bottomrule
\end{tabular}}
\caption{\textbf{Quantitative comparison of motion captioning on HumanML3D}. We use the ground truth texts without pre-processing for linguistic metric calculation.}
\label{tab:caption_performance}
\end{table}

Following other state-of-the-art motion-to-text methods (TM2T~\cite{tm2t}, MotionGPT~\cite{motiongpt}, MotionChain~\cite{motionchain}, and MotionLLM~\cite{motionagent}), we evaluate our motion captioning performance using standard NLP metrics: BLEU~\cite{papineni2002bleu}, ROUGE~\cite{lin2004rouge}, CIDEr~\cite{vedantam2015cider}, and BERTScore~\cite{zhang2019bertscore}. For fair comparison, we adopt Motion-Agent's evaluation approach, using unprocessed ground truth text that ignores tense and plural variations.

As demonstrated in Tab.~\ref{tab:caption_performance}, our motion video caption model \textbf{MVC} (implementation details in Appendix.~\ref{app_subsec:mvs_details}) achieves superior performance in Bleu and Cider metrics, indicating its ability to generate precise and accurate motion descriptions. 
\subsection{Experiments on Challenging Prompts}
\subsubsection{Setup}
To evaluate performance on complex motions, we conducted a user study comparing whole-pipeline \framework~against state-of-the-art open-sourced approaches: MoMask \cite{momask}, ReMoDiffuse \cite{remodiffuse}, and FineMoGen \cite{finemogen}. We designed 40 challenging prompts featuring long, context-rich, spatially compositional motion descriptions (detailed in Appendix.~\ref{app_subsec:challenging_prompts}). The study involved 54 participants evaluating motion sequences across multiple test cases, scoring both motion quality and text-prompt alignment.

We also introduce a novel Motion Alignment Score (MAS) metric, which measures video-text embedding similarity using InternVideo2 \cite{wang2024internvideo2}. This metric enables evaluation of any motion with minimal samples by comparing embeddings from the video encoder (for rendered motion) and text encoder (for prompts). Detailed MAS information is provided in the Appendix.~\ref{app_subsec:mas}.

\subsubsection{Results}
\begin{figure*}[ht]
    \centering
    \includegraphics[width=0.9\textwidth]{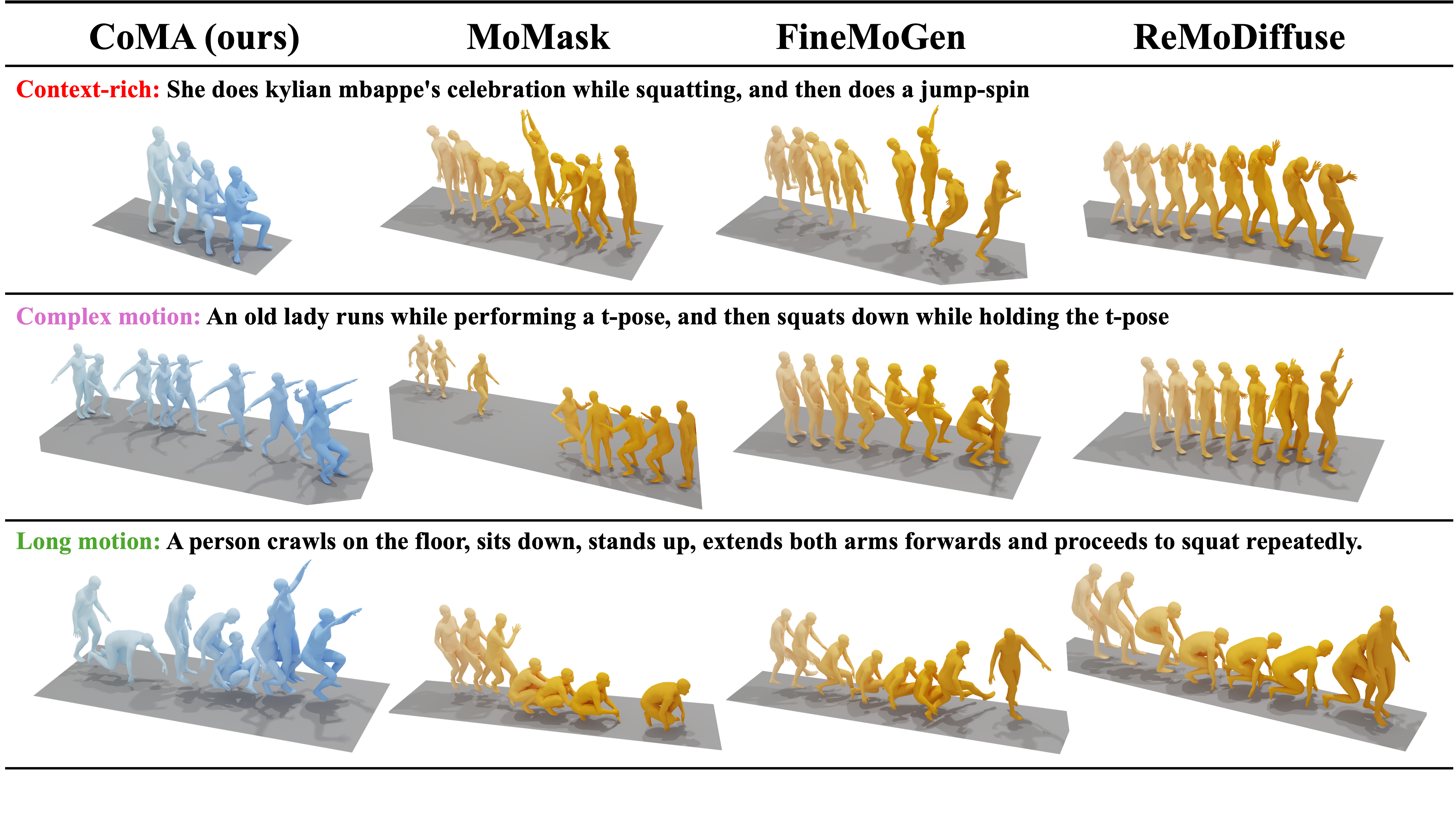}
    \vspace{-2em}
    \caption{A qualitative comparison between \framework\ and state-of-the-art models on three challenging motion tasks: long, complex, and context-rich prompt generation}
    \label{fig:coma_generations}
\end{figure*}
\begin{figure}[ht]
    \centering
    \includegraphics[width=0.9\columnwidth]{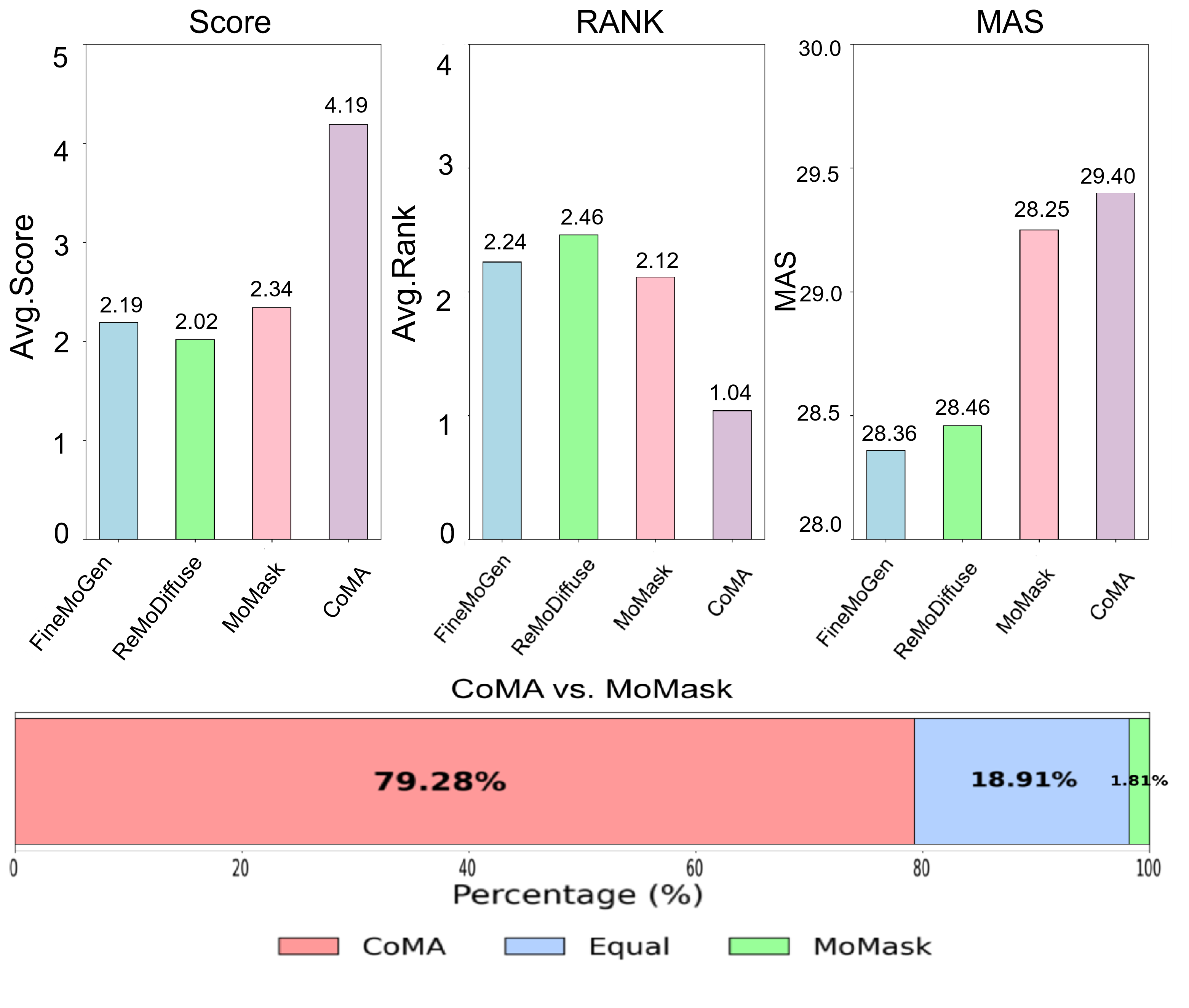}
    \vspace{-1em}
    \caption{User Study Results.}
    \label{fig:user_study}
\end{figure}
Fig.~\ref{fig:user_study} presents comprehensive evaluation results across average score, ranking, and MAS metrics. \framework~consistently outperforms existing approaches across all criteria. In direct comparison with MoMask \cite{momask}, the second-best performer, our method shows superior capability in complex motion generation. Notably, the MAS score improves from 28.61 for first-round generation to 29.40 after editing, highlighting the importance of our iterative refinement pipeline. Visual comparisons in Fig.~\ref{fig:coma_generations} demonstrate our method's significant advantages in generating context-rich, complex, and extended motion sequences.
\subsection{Ablation Study}
The spatial-aware design of the SPAM provides the foundation for the entire pipeline of CoMA. This section ablates the structures of VQVAE and Transformer.

\noindent\textbf{VQVAE Structure}
Our Spatially-Aware VQVAE integrates body parts through a whole-body decoder, which learns to combine them based on the latent vector sequence. MDM\cite{mdm} directly manipulates raw HumanML3D data, but this approach can result in disjointed motion generations. Tab.~\ref{tab:vqablation} highlights the benefits of the whole-body decoder in generation tasks. Direct manipulation of raw data is also unsuitable for motion editing, as it ignores the relationships between body parts. A visual example is provided in the Appendix.~\ref{app_subsec:vqvae_ablation}.
\begin{table}[thb]
    \centering
    \scalebox{0.8}{ % Adjust the scaling factor as needed
        \begin{tabular}{l c c c c}
        \toprule
        \multirow{2}{*}{Methods} & \multicolumn{3}{c}{R Precision$\uparrow$} & \multirow{2}{*}{FID$\downarrow$} \\
        \cline{2-4}
        ~ & Top 1 & Top 2 & Top 3 \\
        \midrule
        WB-D & \best{0.509}{.004} & \best{0.700}{.004} & \secondbest{0.793}{.003} & \best{0.027}{.000} \\
        4S-D & \best{0.509}{.002} & \secondbest{0.698}{.003} & \best{0.794}{.001} & \secondbest{0.046}{.000} \\
        \bottomrule
        \end{tabular}
        }
    \caption{\textbf{Quantitative evaluation on the test split of HumanML3D}. WB-D stands for whole body decoder and 4S-D for 4 separate decoders.}
    \label{tab:vqablation}
\end{table}

\noindent\textbf{Transformer Structure}
The Spatial-Aware Transformer ensures both temporal and spatial consistency in the generated motion, using temporal attention for tokens in different frames and spatial attention for tokens within the same frame. ~\cref{tab:transablation} demonstrates the effectiveness of spatial attention. While full attention performs similarly to our Spatially-Aware Transformer in generation tasks, it lacks precise control over individual body parts and is unsuitable for fine-grained text inputs generated by large language models.

\begin{table}[thb]
    \centering
    \scalebox{0.75}{ % Adjust the scaling factor as needed
        \begin{tabular}{l c c c c}
        \toprule
        \multirow{2}{*}{Methods} & \multicolumn{3}{c}{R Precision$\uparrow$} & \multirow{2}{*}{FID$\downarrow$} \\
        \cline{2-4}
        ~ & Top 1 & Top 2 & Top 3 \\
        \midrule
        Full Attention & \secondbest{0.512}{.002} & \best{0.703}{.006} & \secondbest{0.796}{.005} & \secondbest{0.162}{.009} \\
        SPAM wo spatial & \thirdbest{0.501}{.004} & \thirdbest{0.688}{.004} & \thirdbest{0.791}{.002} & \thirdbest{0.269}{.009} \\
        \bottomrule
        SPAM & \best{0.515}{.007} & \secondbest{0.702}{.001} & \best{0.798}{.004} & \best{0.125}{.008} \\
        \bottomrule
        \end{tabular}
    }
    \caption{\textbf{Quantitative evaluation on the test split of HumanML3D.} SPAM wo spatial refers to the SPAM without spatial attention, Full Attention refers to flattening the latent vector sequence and utilizing full attention.}
    \label{tab:transablation}
\end{table}

\section{Conclusion}

We proposed the multi-modal based, compositional human motion generation framework \framework~to refine complex human motion generations from textual descriptions. With four multi-modal agents powered by a LLM, VLM and a spatially-aware generative motion model, our framework (through \model) performs highly on both standard and complex motion generation quantitative benchmarks. \framework~is capable of longer generations, text-driven editing, motion composition and self-correction, introducing a higher-degree of understanding of the motion domain, leading to higher-quality generations.

\appendix
\section{Experiments on New HumanML3D Split}
\label{sec:complex}
\subsection{Setup}
We developed a new split of the HumanML3D \cite{humanml3d} dataset to specifically evaluate complex motion generation capabilities. Unlike the standard split, which does not differentiate between simple and complex motions, our split deliberately assigns all complex motions to the test set. The resulting distribution includes 20,108 training samples, 2,932 validation samples, and 2,172 test samples. This reorganization shifts the evaluation focus from general performance to specifically measuring how well models can generalize from simple motions to generate more complex sequences.

\begin{table}[thb]
    \centering
    \scalebox{0.8}{ % Adjust the scaling factor as needed
        \begin{tabular}{l c c c c}
        \toprule
        \multirow{2}{*}{Methods} & \multicolumn{3}{c}{R Precision$\uparrow$} & \multirow{2}{*}{FID$\downarrow$} \\
        \cline{2-4}
        ~ & Top 1 & Top 2 & Top 3 \\
        \midrule
        MDM & \thirdbest{0.174}{.003} & \thirdbest{0.289}{.006} & \thirdbest{0.372}{.006} & \thirdbest{5.898}{.195} \\
        % MMM & \et{0.515}{.002} & \et{0.708}{.002} & \et{0.804}{.002} & \et{0.089}{.005} & \et{2.926}{.007} & \et{1.226}{.035}  \\
        MoMask & \secondbest{0.199}{.003} & \secondbest{0.318}{.004} & \secondbest{0.405}{.004} & \best{2.713}{.058} \\
        \midrule
        \model & \best{0.203}{.003} & \best{0.331}{.004} & \best{0.426}{.004} & \secondbest{2.897}{.067} \\
        \bottomrule
        \end{tabular}
    }
    \caption{{Quantitative evaluation on the complex split of HumanML3D}}
    \label{tab:complex_quantitative_eval}
\end{table}

\subsection{Results}
We compared \model~with leading state-of-the-art methods (MDM, MoMask, and MMM) on this complex motion split, with results shown in Tab.~\ref{tab:complex_quantitative_eval}. After training on the original HumanML3D dataset, we fine-tuned our model using detailed text descriptions generated by GPT-4o (as described in Sec.~5.5 of the main paper). \model~delivers superior performance across all R-Precision metrics, while remaining competitive in FID scores, only slightly behind MoMask. We attribute the strong R-Precision results to \model's architecture, which focuses on body part-specific instructions and captures fine-grained movement details from the input text. This design aligns well with R-Precision's goal of measuring semantic correspondence between text descriptions and generated motions.

 \section{User Study}
\label{sec:userstudy}
\subsection{List of Challenging Prompts}
\label{app_subsec:challenging_prompts}
We carefully designed 40 challenging prompts 
(Tab.~\ref{tab:user_study_prompts}) for our user study, considering several key aspects:
\begin{itemize}
    \item \textbf{Context}: Whether the prompts include contextual information. As shown in Tab.~\ref{tab:user_study_prompts}, contexts marked as \orangecheck~are irrelevant to motion (e.g., ``hopeless", ``An away fan") and should be ignored during motion generation. Contexts marked as \greencheck~ are motion-relevant (e.g., ``Spider-Man web shooting move");
    
    \item \textbf{Spatial Composition}: Whether the motion requires coordinated movements of different body parts. For example, ``sit down" is not spatially composite, while ``squat while striking a T-pose" requires coordinated movements;
    
    \item \textbf{Temporal Composition}: Whether the motion consists of multiple segments. Prompts are marked as ``short" for single-segment motions, ``medium" for two segments, and ``long" for more than two segments;
    
    \item \textbf{Explicit Trajectory}: Whether a specific motion path is required. Stationary motions (e.g., ``stand up") and free movements (e.g., ``walk", ``run") don't require explicit trajectories, while specific dynamic motions (e.g., ``Ronaldo's 'Siu' celebration") benefit from predefined trajectories;
    
    \item \textbf{Repeated Motions}: Whether the prompt specifies a number of motion repetitions. As this is a special case of temporal composition, we only include three prompts with this feature in our design.

    \item \textbf{Conversation}: Notably, last two prompts are in question format. This tests our framework's ability to interpret and generate motions from interrogative sentences, demonstrating its potential for natural, conversation-based motion generation interfaces.

\end{itemize}

\subsection{Setup}
We conducted a user study comparing our \framework\ with three state-of-the-art methods: MoMask, ReMoDiffuse, and FineMoGen. These baselines were carefully selected: MoMask and ReMoDiffuse for their strong quantitative performance, and FineMoGen for its focus on complex motion generation. Following the setup of CoMo, we also recruited 54 participants to evaluate generated motion videos. We limited our comparison to four models to maintain a reasonable evaluation load for participants, as each prompt required comparing four different motion videos across 40 different prompts.

For consistent comparison, all frameworks were trained on the standard HumanML3D dataset split. Unlike the previous quantitative evaluations that used only \model, we employed our complete multi-modal pipeline in this study, as the limited number of test prompts made the computational cost of using all agents manageable. To standardize the evaluation process, we limited the self-correction process to two iterations, after which no further edits were made regardless of the Motion Reviewer's assessment.

\subsection{Survey Form}
\begin{figure}[ht]
    \centering
    \includegraphics[width=\columnwidth]{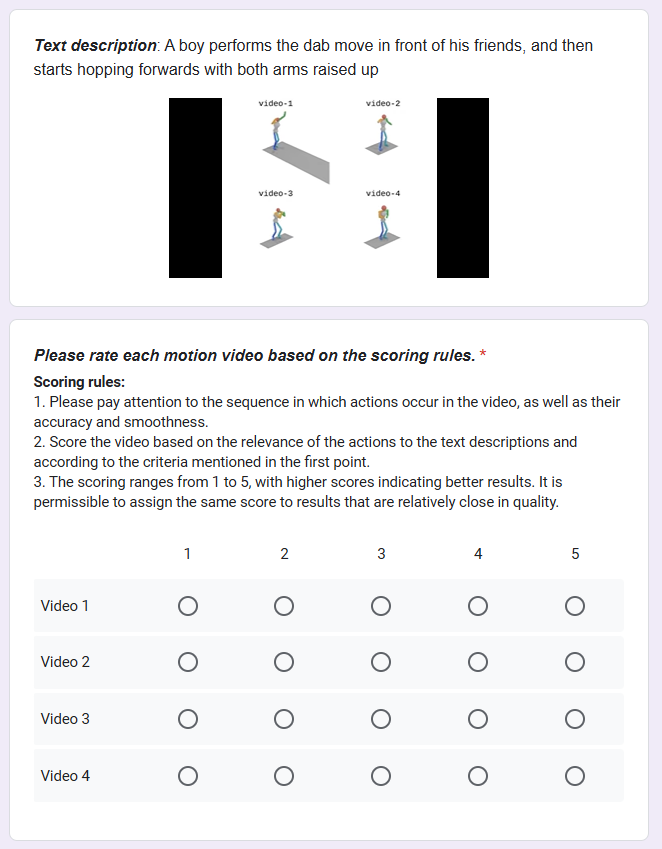}
    \caption{Questionnaire Interface}
    \label{fig:interface}
\end{figure}

As shown in Fig.~\ref{fig:interface}, we designed our questionnaire using Google Forms, with an interface consisting of several key components: text descriptions of the videos, motion sequences, evaluation criteria, and rating matrices. The presented videos include outputs from both our proposed method and other state-of-art approaches, each assigned with a unique identifier to ensure unbiased evaluation. Our evaluation metrics were carefully designed to assess both the intrinsic quality of generated motions and their semantic alignment with the provided text descriptions. To facilitate fair comparison, participants were allowed to assign identical scores to different methods when they judged the quality to be equivalent.

% Then wrap your longtable in strip environment
\onecolumn
\small
\renewcommand{\arraystretch}{1.2}
\begin{longtable}{c|>{\raggedright\arraybackslash}m{6.5cm}|>{\centering\arraybackslash}m{1.6cm}>{\centering\arraybackslash}m{1.9cm}>{\centering\arraybackslash}m{1.9cm}>{\centering\arraybackslash}m{1.8cm}>{\centering\arraybackslash}m{1.2cm}}

\caption{List of motion prompts and their characteristics\label{tab:user_study_prompts}}\\
\toprule
\# & Prompts & Context  & Spatial Composition & Temporal Composition & Explicit Trajectory & Repeated Motions \\
\hline
\endfirsthead

% Header for subsequent pages
\multicolumn{7}{c}{Table \thetable{} continued}\\
\hline
\# & Prompts & Context  & Spatial   Composition & Temporal Composition & Explicit Trajectory & Repeated Motions \\
\hline
\endhead

% Footer for all pages except last
\hline
\multicolumn{7}{r}{\textit{Continued on next page}}\\
\endfoot

% Footer for last page
\hline
\endlastfoot
1 & Thierry Henry's classic celebration during the North London derby against Tottenham in November 2002 at Highbury. & \greencheck & \greencheck & medium & \greencheck & \redx \\
\hline
2 & What is a person's motion before making a penalty kick? & \greencheck & \redx & medium & \greencheck & \redx \\
\hline
3 & A person crawls forward on the floor, transitions to sitting, rises to standing, then with both arms extended forward performs two squats. & \redx & \redx & long & \redx & \greencheck \\
\hline
4 & An angry midfielder performs a slide tackle on another player. & \orangecheck & \redx & short & \redx & \redx \\
\hline
5 & An angry man sits on the court floor looking down, with the right arm to his chest and the left raised upwards to protest a verdict. & \orangecheck & \greencheck & short & \redx & \redx \\
\hline
6 & What is a person's reaction after stubbing their left feet toe? & \redx & \redx & short & \redx & \redx \\
\hline
7 & The Houston Astros manager walked to the field and then performed the Wakanda Forever gesture before the game started. & \orangecheck \greencheck & \redx & short & \redx & \redx \\
\hline
8 & An away fan does the Mbappe celebration to taunt the home team, sits down, stands up and then starts kicking the air. & \orangecheck \greencheck & \greencheck & long & \redx & \redx \\
\hline
9 & A man is running in a zigzag pattern while striking Superman's iconic flying pose. & \greencheck & \greencheck & medium & \greencheck & \redx \\
\hline
10 & The person did the front double bicep pose, switching to a T-pose shortly after, and finally sat down on the floor with the left arm raised and the right arm relaxed. & \greencheck & \greencheck & long & \redx & \redx \\
\hline
11 & A boy performs the dab move in front of his friends, and then starts hopping forwards with both arms raised up. & \redx & \greencheck & medium & \greencheck & \redx \\
\hline
12 & A person does Bruce Lee's classic kicks, and runs forward with right arm extending forward, and trying to avoid sphere obstacles in his way. & \greencheck & \redx & long & \greencheck & \redx \\
\hline
13 & A woman picks up speed from a walk to a run, holding the T-pose. & \redx & \greencheck & medium & \redx & \redx \\
\hline
14 & The girl performs a squat while striking a T-pose. & \redx & \greencheck & short & \redx & \redx \\
\hline
15 & A person performs Black Widow's superhero landing, then slowly stands up. & \greencheck & \greencheck & medium & \redx & \redx \\
\hline
16 & A cowboy does three lasso spins above his head. & \greencheck & \greencheck & short & \redx & \greencheck \\
\hline
17 & He spins while having his right arm upwards and the left arm extended forward. & \redx & \greencheck & short & \redx & \redx \\
\hline
18 & The man sits down, stands up and then does Rocky's victory pose. & \greencheck & \redx & long & \redx & \redx \\
\hline
19 & The boy throws a punch and then a jump spin, afterwards, he starts walking with both arms wide open. & \redx & \greencheck & long & \redx & \redx \\
\hline
20 & A dancer performs a ballet spin. & \greencheck & \greencheck & short & \redx & \redx \\
\hline
21 & A girl did the Home Alone Macaulay Culkin scream pose and then crawled on the floor to find a safe spot. & \greencheck & \greencheck & medium & \redx & \redx \\
\hline
22 & A footballer dives on the field to celebrate a game-winning play. & \greencheck & \greencheck & short & \redx & \redx \\
\hline
23 & A lady falls hopeless on the floor, stands up, raises her right arm and then lowers it. Finally, she jumps with both arms raised. & \orangecheck & \greencheck & long & \redx & \redx \\
\hline
24 & A person does Ronaldo's 'Siu' celebration. & \greencheck & \greencheck & medium & \greencheck & \redx \\
\hline
25 & A person moves in a clockwise circle while alternating between foot taps and hand claps four times in rhythm. & \redx & \greencheck & long & \greencheck & \greencheck \\
\hline
26 & A person runs to center stage and performs a ballet curtsy. & \greencheck & \greencheck & medium & \redx & \redx \\
\hline
27 & A person sits on the floor with hands resting on their knees, then reaches forward with their right arm trying to grab something. & \redx & \greencheck & medium & \redx & \redx \\
\hline
28 & He does the iconic Usain Bolt celebration. & \greencheck & \greencheck & short & \redx & \redx \\
\hline
29 & She does Messi's famous "point to God" celebration. & \greencheck & \greencheck & short & \redx & \redx \\
\hline
30 & She imitates the Hulk's smash stance and then jumps in excitement. & \greencheck & \greencheck & short & \redx & \redx \\
\hline
31 & She performs the iconic Titanic 'flying' pose, followed by a full turn. & \greencheck & \redx & medium & \redx & \redx \\
\hline
32 & Someone walks calmly with their right hand raised in the air, and then sits, and then raises both hands up in the air. Finally, they stand up and start spinning with both arms pointing upwards. & \redx & \redx & long & \redx & \redx \\
\hline
33 & The child does the Spider-Man web shooting move. & \greencheck & \greencheck & short & \redx & \redx \\
\hline
34 & The man does a jump-spin and then a handstand, gets tired and sits down, and then gets up to jump-spin again. & \redx & \greencheck & long & \redx & \redx \\
\hline
35 & The man does the fist of solidarity pose and then squats on the floor with his hands up. Finally, he stands back up again, but now with the right arm extended forward and the left in resting position. & \redx & \redx & long & \redx & \redx \\
\hline
36 & The man does the Tiger Woods fist pump, and then jumps in ecstasy. & \greencheck & \redx & long & \redx & \redx \\
\hline
37 & The mime extends his left arm to the side and the right upwards, as he pretends to be inside a box. & \greencheck & \greencheck & short & \redx & \redx \\
\hline
38 & The soccer player covers their ears in a match-winning goal celebration. & \orangecheck & \redx & short & \redx & \redx \\
\hline
39 & The woman does a fist of solidarity, and afterwards starts running while still holding her right fist up. & \redx & \greencheck & medium & \redx & \redx \\
\hline
40 & Mid-jump kick, the boy loses control and topples over. & \redx & \greencheck & medium & \redx & \redx \\
\bottomrule
\end{longtable}
\twocolumn

\begin{figure*}[ht]
    \centering
    \includegraphics[width=\textwidth]{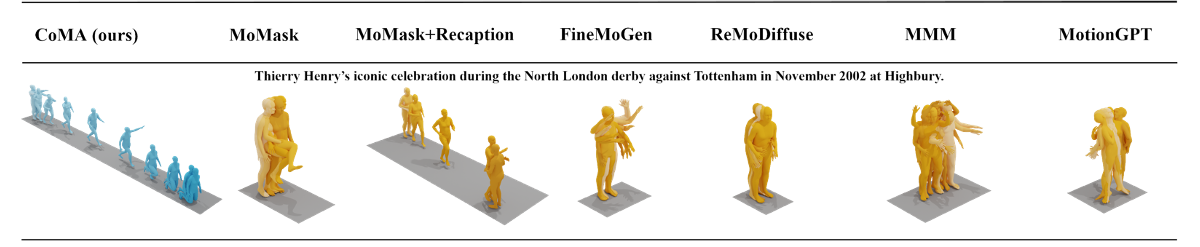}
    \caption{A qualitative comparison between \framework\ and state-of-the-art models in one user study example.}
    \label{fig:user_study_supp}
\end{figure*}

\subsection{Zoom in One Example}
We demonstrate our CoMA workflow using the example shown in Fig.~2 of the main paper. The original motion description \textbf{\emph{``Thierry Henry's classic celebration during the North London derby against Tottenham in November 2002 at Highbury''}} describes a football player's celebratory actions after scoring a goal. 

The \textbf{Task Planner} first transforms this context-rich description into an explicit motion sequence: ``A person sprints with their arms extended over a long distance, then slides on their knees in a straight line.'' Given the temporal complexity, this sequence is decomposed into two segments:
\begin{itemize}
   \item \textbf{Motion Segment \#1}: ``A person sprints with their arms extended over a long distance''
   \item \textbf{Motion Segment \#2}: ``A person slides on their knees in a straight line''
\end{itemize}

To handle the spatial complexity within each segment, the \textbf{Task Planner} further decomposes them into atomic components:
\begin{itemize}
   \item For Segment \#1:
       \begin{itemize}
           \item \textbf{Base Motion}: ``A person runs quickly''
           \item \textbf{Local Edit}: ``The person extends both arms forward''
           \item \textbf{Trajectory}: None (implicit in base motion)
       \end{itemize}
   \item For Segment \#2:
       \begin{itemize}
           \item \textbf{Base Motion}: ``A person kneels down''
           \item \textbf{Local Edit}: None
           \item \textbf{Trajectory}: ``straight line'' (to capture sliding motion)
       \end{itemize}
\end{itemize}

In the generation phase, the \textbf{Motion Generator} creates initial sequences based on the base motions. For Segment \#1, the \textbf{Motion Editor} refines the running motion by adjusting the arm positions, producing Motion Sequence \#1. For Segment \#2, the base motion is generated directly as Motion Sequence \#2.

The \textbf{Motion Reviewer} then evaluates each sequence through video rendering and analysis using our instruction-tuned VideoChat2 model. For Motion Sequence \#1, the generated caption ``A person runs forward with their arms extended'' aligns well with the intended motion when compared using GPT-4o. However, Motion Sequence \#2's caption ``A person sits down'' reveals significant discrepancies. GPT-4o analysis identifies necessary adjustments: the lower body needs to adopt a kneeling position, and the arms require repositioning.

During the refinement phase, the \textbf{Motion Editor} implements identified corrections. While Motion Sequence \#1 requires no changes, Motion Sequence \#2 undergoes adjustments to both arm positions and lower body posture. The \textbf{Trajectory Editor} then enforces the straight-line constraint specified for Segment \#2.

Finally, the \textbf{Motion Generator} combines both sequences into a continuous motion, successfully recreating Thierry Henry's iconic celebration. To validate our approach, we conducted comparative experiments against state-of-the-art baselines, including MoMask with both original and recaptioned prompts. 

As illustrated in Fig.~\ref{fig:user_study_supp}, our method demonstrates superior accuracy and motion fluidity. This example illustrates that our method, through the comprehensive utilization of tools such as task planners, trajectory control, and motion reviewers, generates motions that more closely align with the original prompts and exhibit finer details. In contrast, the motions produced by other methods demonstrate insufficient understanding of the prompts themselves. Furthermore, even when Momask employs a recaption version of the prompt, its model still cannot match our capability in generating complex and content-rich motions.
\section{Task Planner}
\label{app_sec:task_planner}
\subsection{Rewrite Prompts}

The task planner agent's assigned tool is prompt rewriting. Such mechanism has at its core three different prompts, that each play their part in the overall refinement of the initial user input. In the following subsections, we will detail each one in greater depth.

\subsubsection{Text Recaption Prompt}

This prompt's design revolves around analyzing the motion description to infer detailed characteristics of the body’s posture and movements, while restricting GPT-4o's writing with a set of motion-describing words that are known to be present in the training data. The first part of the prompt extracts explicit motion dynamics from potentially vague or complex descriptions, as well as context-reliant terms that are ubiquitous in popular culture but not present in the data. The vocabulary restriction imposed directs the language model to, in the same reasoning step, produce a textual description that closely aligns with our intended goal of providing the clearest and most objective prompt possible.

\begin{tcolorbox}[
    breakable,
    colback=white,
    colframe=black,
    boxrule=0.5pt,
    left=5pt,
    right=5pt,
    top=5pt,
    bottom=5pt
]
\begin{minted}[
    fontsize=\scriptsize,
    breaklines=true,
    breaksymbolleft={},
    breaksymbolright={},
    tabsize=4,
    escapeinside=||
]{text}
You are a reasoning and action model. Your task is to accurately infer the posture and dynamic details of the human body based on a text description of motion, ensuring no information is lost. Your reasoning process involves understanding the motion, synthesizing the overall posture, and providing a detailed description of the movement of the arms, legs, and other body parts. Even if the input motion description is complex, you must strive to present each detail fully without simplifying any actions, ensuring no information is omitted.

The sentence should start with " A person ... " and should be easy to understand and should ONLY use words from the words list below. A word can be used as long as it is mentioned in the **words_list**, regardless of its form. For example, if "walking" is in the **words_list**, then "walks" can also be used.

 |\textbf{\textcolor{Plum}{<words\_list>}}|
    |\textbf{\textcolor{RoyalBlue}{words = [}}|
    |\textbf{\textcolor{RoyalBlue}{'the', 'to', 'a', 'then', 'their', 'right', 'his',}}| 
    |\textbf{\textcolor{RoyalBlue}{'forward', 'with', 'walks', 'left', 'in', 'is', 'up', 'arms', 'back', 'down',}}| 
    |\textbf{\textcolor{RoyalBlue}{'on', 'something', 'steps', 'walking', 'side', 'arm', 'both', 'around', 'it', 'of', 'while', ...,}}| 
    |\textbf{\textcolor{RoyalBlue}{'marches', 'pretends', 'stirring', 'mixing', 'washes', 'pulling', 'also', 'style', 'shrugs', 'wipe', 'hed', 'forearms', 'limping']}}|
|\textbf{\textcolor{Plum}{</words\_list>}}|

Here is an example:

- |\textbf{Motion description}|: The worried contractor walks in a hurry.
- |\textbf{Reasoning 1}|: The description is of a person walking in a hurry.
- |\textbf{Reasoning 2}|: How does a person typically look when they are walking hurriedly? What are the main characteristics of the body during this action?
- |\textbf{Reasoning 3}|: Walking in a hurry means an accelerated pace, with the arms swinging faster, the legs moving quickly, and the body slightly leaning forward.
- |\textbf{Output}|: A person walks in a hurry, with arms swinging faster, quickened steps, and a slight forward lean.

Now do this for the following intput:
|\textcolor{Orange}{\textbf{<Input>}}|
{input_prompt}
|\textcolor{Orange}{\textbf{</Input>}}|
\end{minted}
\end{tcolorbox}

\subsection{Temporal Segment Prompt}

The Temporal Segment Prompt was incorporated in this pipeline to address one of the observed shortcomings of state-of-the-art models, which is prompts that describe a longer motion. Due to the 77 token limit of the CLIP encoder and the overall absence of longer duration motions in HumanML3D, we added this step as a means to further split the output from the Text Recaption prompt into individual temporal segments, straying away from long convoluted sentences and ensuring even more encoder-friendly text. The thought process behind such approach is that, aware of current generation capabilities given a short motion caption with clear description, we sought to infuse such traits in longer prompts, thus making them a concatenation of short motion captions that are known to be understood. 
 
\begin{tcolorbox}[
    breakable,
    colback=white,
    colframe=black,
    boxrule=0.5pt,
    left=5pt,
    right=5pt,
    top=5pt,
    bottom=5pt
]
\begin{minted}[
    fontsize=\scriptsize,
    breaklines=true,
    breaksymbolleft={},
    breaksymbolright={},
    escapeinside=@@,
    tabsize=4
]{text}
The action 'original_action: {original_action}' may require detailed control over specific body parts. 
Please evaluate the action and think carefully about how the movement breaks down into smaller, distinct actions. 
Each step should represent a single, concrete movement without including states or transitional descriptions or stationary motion or pose.
Each step should represent a single, concrete movement without including states or transitional descriptions or stationary motion or pose.

After thinking, provide a structured list of the steps involved in performing this action.

@\textcolor{Orange}{\textbf{<Input>}}@
{input_prompt}
@\textcolor{Orange}{\textbf{</Input>}}@

- Focus on describing the dynamic movement.
- Highlight the necessary coordination between body parts.
- Emphasize the importance of actions: Each step must include key movement details, avoiding redundancy or state descriptions.
- Ensure each step represents a distinct action rather than an intermediate state.
- Streamline the steps: Merge steps as much as possible, ensuring each step contains actual dynamic movements rather than empty descriptions.
- Do not include any description of facial expressions or emotions.
- Focus solely on the action and movement itself.

@\textcolor{Black}{\textbf{The number of steps should be 1, 2, 3, or 4, depending on the TEMPORAL complexity of the action. Do not use too many steps if the action is simple. 2~3 steps are usually enough.}}@

@\textcolor{Black}{\textbf{For each step, use the words 'The man...' or 'The person's ...(body part)' to describe the action.}}@
@\textcolor{Black}{\textbf{Ensure the explanation follows this structure:}}@
@\textcolor{Black}{\textbf{step1: The ...}}@
@\textcolor{Black}{\textbf{step2: The ...}}@
@\textcolor{Black}{\textbf{...}}@


Pay attention to ensure the format is strictly adhered to, as I will break it down according to this structure:
@\textcolor{Plum}{\textbf{<code>}}@
    # @\textcolor{Black}{\textbf{Clean the input sequence\_explanation}}@
    sequence_explanation = sequence_explanation.strip()

    # @\textcolor{Black}{\textbf{Use a regular expression to match all steps and their corresponding descriptions}}@
    # @\textcolor{Black}{\textbf{Pattern explanation:}}@
    # @\textcolor{Black}{\textbf{ - (?m): Multiline mode, enabling \^ and \$ to match the start and end of each line}}@
    # @\textcolor{Black}{\textbf{ - step\textbackslash\text{d}+: Matches step labels, such as step1:}}@
    # @\textcolor{Black}{\textbf{ - \textbackslash\text{s}*: Matches any whitespace characters following the label}}@
    # @\textcolor{Black}{\textbf{ - (.*?)(?=(\textbackslash\text{nstep}\textbackslash\text{d}+:)|\$): Non-greedily matches the description content until the next step label or the end of the string}}@
    pattern = r'(?m)^step\d+:\s*(.*?)(?=(\nstep\d+:)|$)'
    matches = re.findall(pattern, sequence_explanation, re.DOTALL)
    
    result = []
    for match in matches:
        step_description = match[0].strip()
        if step_description:
            step_json = {{
                "prompt": step_description,
                "original prompt": action
            }}
            result.append(step_json)

    return result
@\textcolor{Plum}{\textbf{</code>}}@
\end{minted}
\end{tcolorbox}
 
\subsection{Task Decomposition Prompt}
 
As the last step for the Task Planner's role, we input into GPT-4o the processed outputs from the previous prompt along with our Task Decomposition prompt. For each individual temporal segment, its goal is to convert such caption into base (global) motion and local body part edit tasks, doing so with two distinct prompts: one to extract the base motion, the Base Motion prompt, and another to identify local limb movements as edits, the Local Edit prompt. Working in conjunction, these ensure a clear and systematic decomposition of actions, and generate the final language processing part prior to actually generating the motion.

\subsubsection{Base Motion Prompt}

The Base Motion prompt guides GPT-4o to identify the primary, global movement from the action description, whilst excluding specific limb movements.

\begin{tcolorbox}[
    breakable,
    colback=white,
    colframe=black,
    boxrule=0.5pt,
    left=5pt,
    right=5pt,
    top=5pt,
    bottom=5pt
]
\begin{minted}[
    fontsize=\scriptsize,
    breaklines=true,
    breaksymbolleft={},
    breaksymbolright={},
    tabsize=4,
    escapeinside=@@
]{text}
You are tasked with analyzing the following action description and extracting the base (global) motion component.

@\textcolor{Orange}{\textbf{<Input>}}@
{input_prompt}
@\textcolor{Orange}{\textbf{</Input>}}@

**@\textbf{Definition of Base Motion}@:**
- The Base Motion refers to the primary, overall movement of the entire body.
**@\textbf{Requirements of Base Motion}@:**
- It encompasses the general action without considering specific movements of individual body parts.
- The Base Motion should include head movements and global trajectories but exclude specific movements of the limbs (arms, legs).
- The Base Motion should be simple and clear; clear is important. avoiding use abstract or complex words.
- Do not include any reasoning, explanations, or additional commentary. Use precise and unambiguous language.



- Focus solely on the primary, overall movement, including head movements and general trajectories.
- Exclude any specific movements of the limbs (arms, legs).
- The base motion description should be concise and clear, ideally in one sentence.
- Use precise and unambiguous language.
- Do not include any reasoning, explanations, or additional commentary.

\end{minted}
\end{tcolorbox}

\subsubsection{Local Edit Prompt}

The Local Edit prompt then extracts detailed movements of specific body parts using the Base Motion prompt's output as its starting point. If trajectory data is available in the Base Motion prompt, the Trajectory Editor agent will take over using the result of the Local Edit prompt, otherwise such output will be forwarded to the Motion Generator agent.

\begin{tcolorbox}[
    breakable,
    colback=white,
    colframe=black,
    boxrule=0.5pt,
    left=5pt,
    right=5pt,
    top=5pt,
    bottom=5pt
]
\begin{minted}[
    fontsize=\scriptsize,
    breaklines=true,
    breaksymbolleft={},
    breaksymbolright={},
    escapeinside=@@,
    tabsize=4
]{text}
You are tasked with analyzing the differences between the action description and the base motion to extract local body part movements that need to be applied as edits.

**@\textbf{Definition of Local Edits}@:**
- Local edits refer to specific movements of individual body parts (arms, legs) that occur simultaneously with the base motion.
- These are detailed actions that modify the base motion.

@\textcolor{Orange}{\textbf{<Input>}}@
{input_prompt}
@\textcolor{Orange}{\textbf{</Input>}}@

- Identify all specific movements of the following body parts:
  - "left arm"
  - "right arm"
  - "left leg"
  - "right leg"
- For each body part, describe its movement concisely and specifically in the format:
  "A person's [body part] [action]". OR "A person [action]". 
- For body parts without specific movements, the description should be "none".
- Use clear and unambiguous language.
- Do not include any reasoning, explanations, or additional commentary.
- Include all specified body parts in the output.
- Output only the JSON-formatted local edits.

Provide the local edits in the following JSON format, enclosed in <LOCAL_EDITS_JSON> tags:


@\textcolor{Plum}{\textbf{<LOCAL\_EDITS\_JSON>}}@
[
  {{
    "body part": "left arm",
    "description": "[specific movement or 'none']"
  }},
  {{
    "body part": "right arm",
    "description": "[specific movement or 'none']"
  }},
  {{
    "body part": "left leg",
    "description": "[specific movement or 'none']"
  }},
  {{
    "body part": "right leg",
    "description": "[specific movement or 'none']"
  }}
]
@\textcolor{Plum}{\textbf{</LOCAL\_EDITS\_JSON>}}@
\end{minted}
\end{tcolorbox}

% \subsubsection{Process Workflow}

% The motion decomposition process involves the following steps:
% \begin{enumerate}
%     \item Analyze the input description using the Base Motion prompt to generate a concise, global description of the primary movement.
%     \item The Local Edit prompt is applied to identify specific limb movements, which are expressed in a structured JSON format for clarity and consistency.
% \end{enumerate}

\section{Trajectory Editor}
\label{app_sec:trajectory_editor}
\subsection{Trajectory Generation Prompts}

We enforce a Chain-Of-Thought prompting strategy to generate continuous trajectories representing specified shapes or paths. These prompts are designed to guide a language model in producing mathematical functions that define such trajectories.

\begin{tcolorbox}[
    breakable,
    colback=white,
    colframe=black,
    boxrule=0.5pt,
    left=5pt,
    right=5pt,
    top=5pt,
    bottom=5pt
]
\begin{minted}[
    fontsize=\scriptsize,
    breaklines=true,
    breaksymbolleft={},
    breaksymbolright={},
    escapeinside=@@,
    tabsize=4
]{text}
**@\textbf{Task}@:** Draw a continuous trajectory to represent a specified curve/line/shape(trajectory) of a person, according to the given input.

@\textcolor{Orange}{\textbf{<Input>}}@
{input_prompt}
@\textcolor{Orange}{\textbf{</Input>}}@

**@\textbf{YOUR OUTPUT SHOULD CONTAIN}@:**

@1@. **@\textbf{Closed or Open Trajectory Decision}@:** Decide if the trajectory is closed or open based on the description. For example, if it's a geometric figure or involves "walking around," it's likely closed. If it's a path like the letter 'S', 'L', etc., it's open. So, avoid using a closed trajectory for an open path like S, a common error is to make it like shape 8.

@\textbf{2}@. **@\textbf{Extract the Trajectory Using Fixed Format Breakdown}@: (ONLY DO WHEN Trajectoy is complex or vague. If it's simple, you can skip this step)** Break down the action description into simple, precise steps. Use a fixed format to describe the movement (e.g., "Walk forward for 5 meters, then turn 90 degrees right"). This helps in extracting the trajectory.

@\textbf{2.1}@ @\textbf{Avoid overcomplicating the movement.Keep it accurate and straightforward}@.

@\textbf{3}@. **@\textbf{Trajectory Analysis}@:** Analyze the described trajectory before writing the code. Consider overlapping parts where necessary (it's not normal curve, it's trajectory. A man's trajectory can overlap). The parameter `t` in `shape_curve(t)` may represent time in some cases.

**@\textbf{Note}@:**: Your understanding of clock directions might be different from mine, so here's a quick reference:
12 o'clock: Straight ahead
3 o'clock: Directly to your right
6 o'clock: Directly behind you
9 o'clock: Directly to your left
1-2 o'clock: Slightly to the right front
10-11 o'clock: Slightly to the left front
4-5 o'clock: Slightly to the right back
7-8 o'clock: Slightly to the left back

**@\textbf{Note}@:**: Whether it's to the right, left, or any clock direction, it's always referenced from the perspective of the person walking this trajectory, not from the image's perspective.

**@\textbf{Note}@:** Clock directions are always referenced from the perspective of the person performing the trajectory.
Ensure that both x and y coordinates change uniformly over time (`t`). This means the trajectory should reflect a consistent speed of movement.
To ensure no "instant jumps" in the generated trajectory, specify that the trajectory function must have smooth transitions between segments. Emphasize continuity, meaning each segment's start must align with the previous segment's end, avoiding abrupt shifts. Additionally, ensure uniform speed across the entire range of `t`, with x and y coordinates changing evenly over time.

Emphasize once again: To ensure a smooth transition, you need to adjust the formulas for each segment so that they start from the endpoint coordinates of the previous segment, rather than independently redefining ( x ) and ( y ). If there are multiple segments, the starting and ending coordinates for each segment should be clearly marked in the comments as **start_x, start_y, end_x, end_y**.
Emphasize once again: To ensure a smooth transition, you need to adjust the formulas for each segment so that they start from the endpoint coordinates of the previous segment, rather than independently redefining \( x \) and \( y \). If there are multiple segments, the starting and ending coordinates for each segment should be clearly marked in the comments as **start_x, start_y, end_x, end_y**.

@\textbf{4}@. **@\textbf{Mathematical Functions}@:** Present the final code strictly in the form provided below, ensuring it is correct and can run without errors and READY TO USE.

**@\textbf{Code Format}@:**
```python
def shape_curve(t):xz
    ...

    return x, y

# Specify the range of t (it is important)
t_range = (start_value, end_value)
```
Now the input is: "I want to draw a Description = 'placeholder1'. Give me `def shape_curve(t)`."
\end{minted}
\end{tcolorbox}

\vspace{1em}
Trajectory generation prompts contain the following key instructions:
\begin{itemize}
    \item Closed or Open Trajectory Decision: Determine whether the trajectory is closed or open based on the description. For example, geometric figures like circles are closed, while paths like the letter 'S' are open.
    \item Trajectory Analysis: Analyze the trajectory, considering overlapping parts and ensuring consistent movement speed. The model should ensure that the trajectory reflects a uniform speed and smooth transitions without abrupt shifts.
    \item Mathematical Functions: Present the final code in the specified format, ensuring it is correct, executable, and ready to use.
\end{itemize}
\vspace{1em}

Once the trajectory is mathematically defined, it is evaluated by sampling points along the curve, dividing the range of $t$ into 200 evenly spaced steps, due to the common 196 frame cap on generations from state-of-the-art methods; such sampling also ensures adaptability for motions of varying lengths. Each $t$ value is then passed through the function \texttt{shape\_curve(t)} function to generate corresponding $x, y$ ground coordinates.

These points are then converted into velocity vectors by calculating differences between consecutive points. To ensure smooth and uniform motions, the trajectory is resampled using B-spline interpolation, which adjusts point spacing and eliminates irregularities. This output is then normalized to a standard velocity range, calculated from the original motion. The result is a smooth, consistent trajectory, ready for use in simulating realistic movement patterns.

\subsection{Trajectory Mapping}
From the resampled trajectory, we compute two key velocity components: $v_{rot}^{root}$, which defines the frame-to-frame rotational adjustments, and $v_{lin}^{root}$, which determines the character's facing direction and movement speed.

The derived velocity data undergoes processing and scaling. Rotational velocity $v_{rot}$ is scaled to match $\bar{v}$, the standard mean velocity from the original motion, ensuring preservation of motion dynamics. This processed rotational velocity then replaces $v_{rot}^{root}$ in the original motion data $\mathbf{M} \in \mathbb{R}^{T \times 263}$, where $T$ represents sequence length and $v_{rot}^{root}$ corresponds to the first dimension of the 263 features.

Generally, our pipeline maintains the original $v_{lin}^{root} \in \mathbb{R}^{T \times 2}$, which represents root linear velocity in the second and third dimensions of the motion features. 
% However, for specific cases like extended kneeling slides, we adjust the movement speed by updating $v_{lin}^{root}$ accordingly.

\subsection{Zoom in One Example}
\label{app_subsec:zoom_in_example}
Given the prompt \textbf{\emph{``A man walks in the trajectory that forms the shape of a heart''}}, our Trajectory Editor agent performs the following analysis:

\begin{tcolorbox}[
    breakable,
    colback=white,
    colframe=black,
    boxrule=0.5pt,
    left=5pt,
    right=5pt,
    top=5pt,
    bottom=5pt
]
\begin{minted}[
    fontsize=\scriptsize,
    breaklines=true,
    breaksymbolleft={},
    breaksymbolright={},
    escapeinside=@@,
    tabsize=4
]{text}
@\textbf{Model}@: To create a trajectory that forms the shape of a heart, we need a trajectory that outlines the iconic curves and point of a heart. The heart shape can be broken down into two symmetric circular arcs at the top and a pointed end at the bottom.

Given that the heart shape is symmetric and closed, we ensure the trajectory smoothly follows this pattern. Let's start by defining the trajectory:

### @\textbf{Closed or Open Trajectory Decision}@:
The heart shape is a closed trajectory.

### @\textbf{Trajectory Analysis}@:
A heart shape consists of two symmetrical arcs at the top and a point at the bottom. We can mathematically approximate this using parametric equations for a heart shape.

### @\textbf{Mathematical Functions}@:
Let's implement a function for drawing a heart shape trajectory:

```python
import numpy as np

def shape_curve(t):
    # Heart shape parametric equations
    x = 16 * np.sin(t) ** 3
    y = 13 * np.cos(t) - 5 * np.cos(2 * t) - 2 * np.cos(3 * t) - np.cos(4 * t)
    
    return x, y

# Specify the range of t (it is important)
t_range = (0, 2 * np.pi)
```

### @\textbf{Explanation}@:
- The parametric equations for `x` and `y` are chosen to represent the heart shape. These equations generate the characteristic curves and pointed bottom of a heart.
- `t` ranges from `\theta` to `2\pi`, allowing a full cycle of the heart shape to be created smoothly and continuously.

This code is ready to plot a heart-shaped trajectory when fed into a function that evaluates it over the specified `t_range`.

\end{minted}
\end{tcolorbox}

\begin{figure}[ht]
    \centering
    \vspace{-1em}
    \includegraphics[width=0.7\columnwidth]{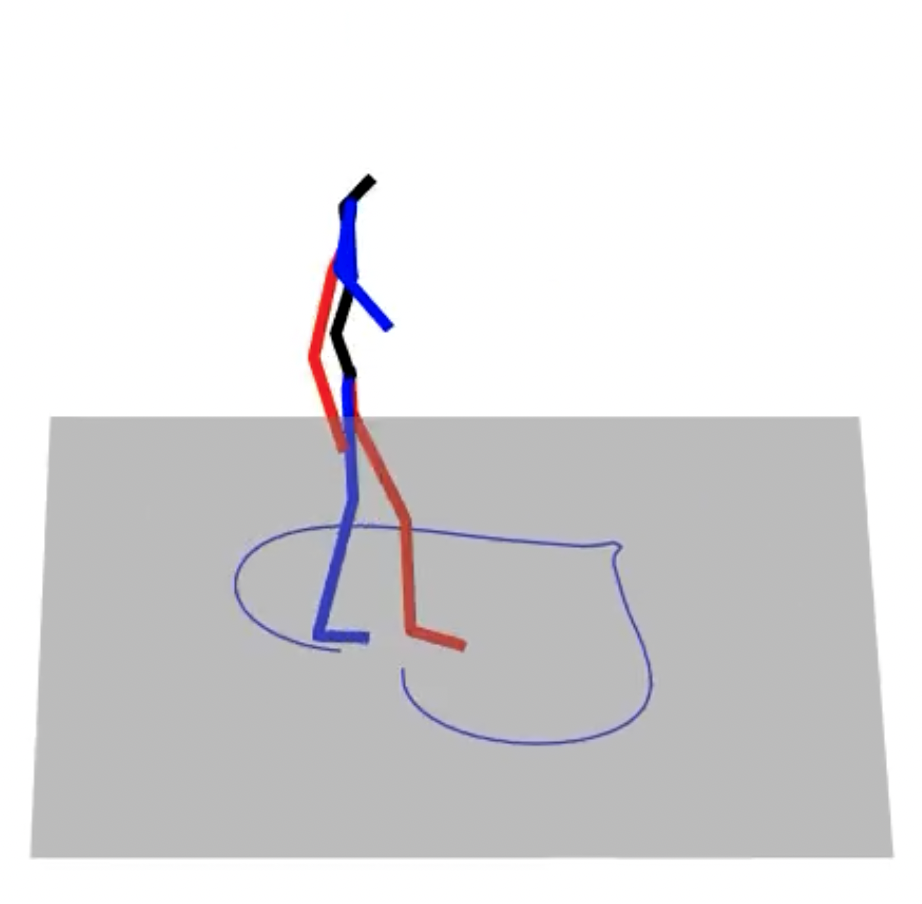}
    \caption{The result of trajectory guidance}
    \label{fig:traj_edit}
\end{figure}

Fig.~\ref{fig:traj_edit} shows the generated heart-shaped walking trajectory.

\section{Motion Generator}
\label{app_sec:motion_generator}

\subsection{Implementation Details}
\label{app_subsec:spam_details}
\model~is implemented in PyTorch. The spatially-aware residual VQ-VAE uses residual blocks for both encoder and decoder with a downscale factor of 4. Each body part's VQ module contains 6 quantization layers, each with 512 codes of 128 dimensions, and we set the quantization dropout ratio to 0.2. Both base and residual transformers comprise 9 transformer layers with 8 attention heads and a latent dimension of 512. We employ a learning rate of 2e-4 with linear warm-up over 2000 iterations, and batch sizes of 512 and 256 for VQ-VAE and transformer training, respectively. During inference, we use Classifier-Free Guidance scales of 4 (base transformer) and 5 (residual transformer), with 10 inference time steps. All experiments involving only \model~can be executed on a single NVIDIA 3090 Ti GPU. The VQ-VAE is trained for 30 epochs using the training split. Meanwhile, the base transformer and residual transformer are trained for 750 and 500 epochs, respectively. 

\subsection{Body Parts Division}
\label{app_subsec:body_parts}
Inspired by \cite{finemogen} , we divided the human body into fine-grained body parts. Specifically, the body is partitioned into four parts: right upper, left upper, right lower, and left lower. The names of the joints included in each part are listed below. Note that there is some overlap between the right upper and left upper bodies, as well as between the right lower and left lower.  
\begin{tcolorbox}[
    breakable,
    colback=white,
    colframe=black,
    boxrule=0.5pt,
    left=5pt,
    right=5pt,
    top=5pt,
    bottom=5pt
]
\begin{minted}[
    fontsize=\scriptsize,
    breaklines=true,
    breaksymbolleft={},
    breaksymbolright={},
    escapeinside=@@,
    tabsize=4
]{text}

@\textbf{Left upper:}@
'left_collar', 'left_shoulder', 'left_elbow', 'left_wrist', 'spine3', 'spine2', 'spine1', 'head', 'neck'

@\textbf{Right upper:}@
'right_collar', 'right_shoulder', 'right_elbow', 'right_wrist', 'spine3', 'spine2', 'spine1', 'head', 'neck'

@\textbf{Left lower:}@
'left_ankle', 'left_foot', 'left_hip', 'pelvis', 'left_knee'

@\textbf{Right lower:}@ 
'right_ankle', 'right_foot', 'right_hip', 'pelvis', 'right_knee'

\end{minted}
\end{tcolorbox}

Another intuitive way to partition the body is to separate the torso, as shown in below, resulting in five parts: right upper, left upper, right lower, left lower, and torso. 

\begin{tcolorbox}[
    breakable,
    colback=white,
    colframe=black,
    boxrule=0.5pt,
    left=5pt,
    right=5pt,
    top=5pt,
    bottom=5pt
]
\begin{minted}[
    fontsize=\scriptsize,
    breaklines=true,
    breaksymbolleft={},
    breaksymbolright={},
    escapeinside=@@,
    tabsize=4
]{text}

@\textbf{Left upper:}@
'left_collar', 'left_shoulder', 'left_elbow', 'left_wrist'

@\textbf{Right upper:}@
'right_collar', 'right_shoulder', 'right_elbow', 'right_wrist'

@\textbf{Left lower:}@
'left_ankle', 'left_foot', 'left_hip', 'left_knee'

@\textbf{Right lower:}@
'right_ankle', 'right_foot', 'right_hip', 'right_knee'

@\textbf{Torso:}@
'spine3', 'spine2', 'spine1', 'head', 'neck', 'pelvis'

\end{minted}
\end{tcolorbox}
 
However, we argue that this partitioning does not align well with the natural language descriptions. Large language models struggle to describe motions involving the torso, leading to lower-quality local text generation for this region. To avoid this issue, we divide the body into four parts without separating the torso, allowing for overlap between the regions. This partitioning approach simplifies the reasoning process for large language models, aligns better with natural language conventions, and enables users to design local text descriptions more easily. 

\subsection{VQVAE Structure}
\label{app_subsec:vqvae_ablation}
Our spatially-aware VQVAE not only segments the body into local parts but also integrates the localized parts into a complete motion. The decoder is responsible for reconstructing the complete motion from the localized tokens. Specifically, we concatenate the token embeddings of the four body parts and input them into the whole-body decoder. The whole-body decoder learns to transform the concatenated embeddings into a complete motion while ensuring the spatial consistency. 

\begin{figure*}[!ht]
    \centering
    \includegraphics[width=0.8\textwidth]{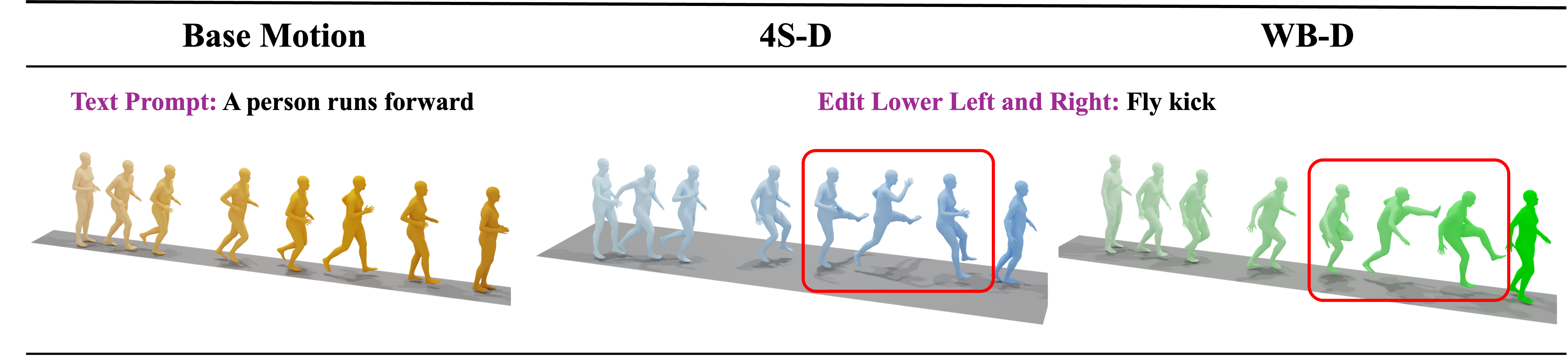}
    \caption{\textbf{Motion editing comparison between separate body-part decoders (4S-D) and whole-body decoder (WB-D).} The whole-body approach demonstrates superior adherence to editing instructions while maintaining physical plausibility.}
    \label{fig:vqvae_ablation}
\end{figure*}

Previous work \cite{mdm} proposed directly manipulating the HumanML3D \cite{humanml3d} data by integrating motions at the raw motion level instead of in the latent space. Based on this, we could train multiple local decoders and then merge the local motions. However, this approach neglects the relationships between different body parts, leading to unnatural results in both generation and editing. Furthermore, it limits the possibility of performing multiple edits on the motion. As shown in Fig.~\ref{fig:vqvae_ablation}, the example highlights how this integration approach restricts the model’s understanding of the spatial structure of the human body and disrupts the balance during editing.

\section{Motion Reviewer}
\label{app_sec:motion_reviewer}
\subsection{Motion Render}
We utilize the rendering tools from the MotionGPT official repository\footnote{\url{https://github.com/OpenMotionLab/MotionGPT}}. Our visualization represents the human body using joints, with different colors assigned to distinct body parts, as illustrated in Fig.~\ref{fig:spam} of the main paper. While this joint-based representation is not photorealistic, the color-coded body parts enhance the ability to distinguish local movements during video captioning.

\subsection{MVS Implementation Details}
\label{app_subsec:mvs_details}
We instruction-tuned VideoChat2 VLM, the core component of our Motion Reviewer agent, using 11,692 training samples from the HumanML3D dataset. To optimize training efficiency, we excluded mirrored samples (prefixed with ``M'') and downsampled each motion to 40 frames. The training process, conducted on an A100 GPU, completed in three days over 20 epochs.

For training, we used a consistent instruction format: ``Describe the motion of the person rendered as a stick figure in the video.'' Each sample consisted of the video input and its corresponding text description as the answer, with an empty question field as the task focused solely on caption generation.

\subsection{Correction Instruction Prompts}
For generating motion correction instruction, we leverage the capabilities of large language models (LLMs). Specifically, we use GPT-4o to compare the motion caption with the corresponding prompt derived from text recaption. The process begins by parsing both the caption and the prompt into individual body part descriptions. GPT-4o then compares the parsed body part descriptions from the caption and the prompt to determine which parts are aligned and which require adjustment.
Finally, based on this comparison, GPT-4o generates precise editing instructions, specifying necessary modifications for the arms and the lower body.

The prompt used for decomposing text into body part descriptions is as follows:
\begin{tcolorbox}[
    breakable,
    colback=white,
    colframe=black,
    boxrule=0.5pt,
    left=5pt,
    right=5pt,
    top=5pt,
    bottom=5pt
]
\begin{minted}[
    fontsize=\scriptsize,
    breaklines=true,
    breaksymbolleft={},
    breaksymbolright={},
    escapeinside=@@,
    tabsize=4
]{text}
Your task is to generate different body parts motion according to a Motion Description. The body parts are right arm, left arm, right leg and left leg.

You only need to output motions of different body parts without any explanation. If some body parts are not mentioned in the Motion Description, you need to deduce those body parts by the Motion Description. Ensure that the motion described is rational and appropriate for the specified body part, aligning with the original motion description. In the final motion description, the body parts must be the subject of the sentence.

### @\textbf{The input format is}@:
    @\textcolor{RoyalBlue}{\textbf{Motion Description: [Insert text here]}}@
    
### @\textbf{The output format is}@:
    @\textcolor{Plum}{\textbf{Right arm: [the final right arm motion description including right arm as the subject.]}}@
    
    @\textcolor{Plum}{\textbf{Left arm: [the final left arm motion description including left arm as the subject.]}}@
    
    @\textcolor{Plum}{\textbf{Right leg: [the final right leg motion description including right leg as the subject.]}}@
    
    @\textcolor{Plum}{\textbf{Left leg: [the final left leg motion description including left leg as the subject.]}}@

@\textcolor{Orange}{\textbf{<Input>}}@
{input_prompt}
@\textcolor{Orange}{\textbf{</Input>}}@
\end{minted}
\end{tcolorbox}

The prompt used for comparing body part descriptions is as follows:
\begin{tcolorbox}[
    breakable,
    colback=white,
    colframe=black,
    boxrule=0.5pt,
    left=5pt,
    right=5pt,
    top=5pt,
    bottom=5pt
]
\begin{minted}[
    fontsize=\scriptsize,
    breaklines=true,
    breaksymbolleft={},
    breaksymbolright={},
    escapeinside=@@,
    tabsize=4
]{text}
You have two groups of motion descriptions stored in dictionaries. Each dictionary contains the following keys: ‘motion’, ‘Right arm’, ‘Left arm’, ‘Right leg’, and ‘Left leg’. The ‘motion’ key describes a person’s overall movement, while the other keys specify the movement of each body part in that motion.

### @\textbf{Your task}@:
	Compare the ‘motion’ in two motion descriptions: ‘motion description1’ (the standard motion) and ‘motion description2’ (the observed motion).
    
	Determine if the ‘motion’ in ‘motion description2’ approximately matches the ‘motion’ in ‘motion description1’. 
    
    if there is a mismatch in a specific body part, you should generate what this body part should do so that it can match ‘motion description1’.

### @\textbf{Guidelines}@:
    Only use 'motion' to do comparision.
	
    ##@\textbf{If there is a mismatch}@:
	   **@\textbf{For left arm mismatches}@:** 
        use the ‘Left arm’ in ‘motion description1’ to help you understand the left arm motion (do not directly use it to generate your answer) and then generate an left arm motion instruction. 
        
        **@\textbf{For right arm mismatches}@:** 
        use the ‘Right arm’ in ‘motion description1’ to help you understand the right arm motion (do not directly use it to generate your answer) and then generate an upper body motion instruction. 
	    
        **@\textbf{For lower body mismatches}@:** 
        use the ‘Right leg’ and ‘Left leg’ motions in ‘motion description1’ to help you understand the upper body motion (do not directly use it to generate your answer) and then generate a lower body motion instruction. The lower bdoy motion must be cohesive and naturely. 
        
    The body part motion is just for reference and help you better understand. Don't directly use the body part motion to generate output. Please start you answer from 'motion' in the motion description1. Remember motion description1 is the standard one! 
	    
    If the 'motion' of two motion description are approximately same, describing a similar motion, both upper body and lower body output None. You don't need to pay attention to the detail of two motion. We only need two motions are approximately same.
       
    **@\textbf{Approximately same}@:** 
    if two specific and corresponding body part do a same action (raise, jump, ...), they are approximately same. You do not need to pay attention to the height of arm raised and how far a peson jump. This is the detail of one action. You do not need to pay attention to the detail of action.
       
    **@\textbf{For example}@:** 
    the first motion that the man is walking clockwise in a circle while holding something up to his ear with his left arm. The second motion that a man with his left arm raised walk clockwise. The person in two motions both walk clockwise and raise their left arm. So these two motions are approximately same.

### @\textbf{Output Requirements}@:
	For mismatched motions, output only the motion instruction for the person’s left arm or right or lower body without explanation. You must use 'a person' as the subject of your output motion for all body parts!!!
	
    For matched motions, simply output “None” for the respective body part.

### @\textbf{Input Format}@:
    @\textcolor{RoyalBlue}{\textbf{Motion Description1: [Insert text here]}}@
    
    @\textcolor{RoyalBlue}{\textbf{Motion Description2: [Insert text here]}}@


### @\textbf{Output Format}@:
    @\textcolor{Plum}{\textbf{Left arm:   [Insert motion or "None"]}}@
    
    @\textcolor{Plum}{\textbf{Right arm:  [Insert motion or "None"]}}@
    
    @\textcolor{Plum}{\textbf{Lower body: [Insert motion or "None"]}}@

@\textcolor{Orange}{\textbf{<Input>}}@
{input_prompt}
@\textcolor{Orange}{\textbf{</Input>}}@
\end{minted}
\end{tcolorbox}

\subsection{Motion Video Alignment (MAS) Score}
\label{app_subsec:mas}
\begin{table}[htp]
    \centering
    \scalebox{0.7}{
    \begin{tabular}{l c c c c c c}
    \toprule
    \multirow{2}{*}{Methods}  & \multicolumn{3}{c}{R Precision$\uparrow$} & \multirow{2}{*}{MM Dist$\downarrow$} & \multirow{2}{*}{MAS$\uparrow$}\\

    \cline{2-4}
    ~ & Top 1 & Top 2 & Top 3 \\
    \midrule

MDM & \et{0.362}{.015} & \thirdbest{0.534}{.008} & \thirdbest{0.648}{.007} & \thirdbest{3.851}{.051} & \thirdmas{31.793} \\
MotionGPT & \thirdbest{0.365}{.007} & \et{0.524}{.006} & \et{0.607}{.008} & \et{4.512}{.047} & 31.538 \\
MoMask & \secondbest{0.409}{.010} & \secondbest{0.588}{.010} & \secondbest{0.700}{.007} & \secondbest{3.689}{.054} & \secondmas{31.929} \\
\midrule
\textbf{SPAM (ours)} & \best{0.435}{.008} & \best{0.597}{.009} & \best{0.703}{.008} & \best{3.665}{.044} & \bestmas{31.961} \\
    \bottomrule
    \end{tabular}
    }

\caption{\textbf{Quantitative evaluation between R Precision, MultiModal Dist and MAS.} $\pm$ indicates a 95\% confidence interval.}
\label{tab:MAS}
\end{table}

We propose a new evaluation metric to assess the quality of motion generation from a video perception perspective. Specifically, we utilize both the Video Encoder and Text Encoder from InternVideo2 to encode video and text inputs, respectively. The \textbf{Motion Alignment Score (MAS)} is calculated as 100 times the cosine similarity between the text embedding and its corresponding video embedding. This allows us to evaluate the quality of any unseen motion outside the training dataset.

To validate the effectiveness of MAS, we randomly selected 320 test samples from the HumanML3D dataset and calculated R-Precision and MMDist metrics. We then compared our model against MoMask, MotionGPT, and MDM. In Tab.~\ref{tab:MAS}, the ranking of MAS scores aligns closely with the rankings of R-Precision (Top2, Top3) and MMDist, demonstrating that MAS effectively evaluates motion generation quality.
\section{Limitations and Future Work}
\label{sec:limitations}
\framework\ has several limitations. First, the inference speed is constrained by motion video rendering in Blender. One potential solution is to replace the video caption model with a motion caption model. However, without a pre-trained large motion-language model, the out-of-domain performance of such a motion caption model remains uncertain.

Second, we have not thoroughly investigated the impact of multiple self-correction iterations. While most generated motions show significant improvement after a single round of VLM-guided correction, in some cases, further refinement can lead to performance degradation. This suggests the need for more robust criteria to determine when to halt the refinement process.

Finally, \framework\ is not trained end-to-end. We are currently exploring the development of a comprehensive human-centric multimodal language model that can seamlessly integrate text, motion, image, video, and sound for generation, editing, and reasoning tasks.  

%%%%%%%%% REFERENCES
{\small
\bibliographystyle{ieeenat_fullname}
\bibliography{main}
}

\end{document}